%% file: ms.tex
\documentclass{article}
\pdfoutput=1
\usepackage[hyperfootnotes=false]{hyperref}
\usepackage[utf8]{inputenc}
\usepackage[table]{xcolor}
\usepackage{nips15submit_e}
\usepackage[square,sort,numbers]{natbib}
\usepackage{caption}

\usepackage{booktabs}
\usepackage{amssymb}
\usepackage{amsmath, amssymb, amsthm}
\usepackage{graphicx}
\usepackage[font=footnotesize,caption=false]{subfig}
\usepackage{algorithm}
\usepackage{algpseudocode}
\usepackage{capt-of}
\usepackage[small,tight]{bibhacks}
\usepackage{chngpage}

\newcommand{\doi}[1]{\textsc{doi}: \href{http://dx.doi.org/#1}{\nolinkurl{#1}}}

\newcommand\blfootnote[1]{%
  \begingroup
  \renewcommand\thefootnote{}\footnote{#1}%
  \addtocounter{footnote}{-1}%
  \endgroup
}

\title{Explainable Black-Box Attacks Against Model-based Authentication}

\author{
Washington Garcia \\
University of Florida \\
\texttt{w.garcia@ufl.edu} \\
\And
Joseph I. Choi \\
University of Florida \\
\texttt{choijoseph007@ufl.edu} \\
\And
Suman K. Adari \\
University of Florida \\
\texttt{sadari@ufl.edu} \\
\And
Somesh Jha \\
University of Wisconsion--Madison \\
\texttt{jha@cs.wisc.edu}
\And 
Kevin R. B. Butler \\
University of Florida \\
\texttt{butler@ufl.edu} \\
}

\nipsfinalcopy 

\begin{document}
	\maketitle
	
	\blfootnote{Pre--print, work in progress.}

	\begin{abstract}
		\input{abstract}
	\end{abstract}

	\input{intro}

	\input{background}

	\input{design}
	\input{eval}

	\input{discuss}
	\input{relwork}

	\input{conc}

	\bibliographystyle{abbrv}
	\bibliography{literatur,choi}
\end{document}

%% file: abstract.tex
Establishing unique identities for both humans and end systems has been an active research problem in the security community, giving rise to innovative machine learning-based authentication techniques. Although such techniques offer an automated method to establish identity, they have not been vetted against sophisticated attacks that target their core machine learning technique. This paper demonstrates that mimicking the unique signatures generated by host fingerprinting and biometric authentication systems is possible. We expose the ineffectiveness of underlying machine learning classification models by constructing a blind attack based around the query synthesis framework and utilizing Explainable--AI (XAI) techniques. We launch an attack in under 130 queries on a state-of-the-art face authentication system, and under 100 queries on a host authentication system. We examine how these attacks can be defended against and explore their limitations. XAI provides an effective means for adversaries to infer decision boundaries and provides a new way forward in constructing attacks against systems using machine learning models for authentication. 

%% file: intro.tex
\section{Introduction}
\label{sec:intro}

Authenticating subjects in a system is essential for establishing trust. While
authentication is often performed with 
traditional credential or user-based mechanisms,
machine learning (ML) is being increasingly used for authentication.
By reformulating authentication as a classification problem, both humans and
machines can act as subjects and be authenticated with high accuracy and
minimal false positives~\cite{Bates2014,
  Radhakrishnan2015,Kohno2005,gaussianface,Hong1997,Bigun2005}. The underlying
goal of these authentication systems is to use unique information owned by the user to attest a claimed identity. The modality of this information has changed from passphrases and PIN numbers to fuzzy features such as facial features~\cite{Bigun2005} and end-system behavior (e.g., key strokes~\cite{Mondal2017}, hardware information~\cite{Bratus}, and device usage~\cite{Mondal2017,Jakobsson,DeLuca}). 

However, these authentication techniques tend to only consider naive spoofing attacks that replicate features using domain knowledge, rather than targeting the underlying model's decision boundary. The field of adversarial machine learning (AML) has uncovered several milestone attacks against state-of-the-art ML models~\cite{Papernot2015, Papernot2016, Fredrikson2015}. A limitation of these proposed attacks is the context in which they are built; their explicit goal is breaking classification tasks, rather than authentication tasks. 
Some of the key differences that distinguish authentication attacks from recently proposed classification attacks are:

\noindent
\textbf{Information returned from the authentication system is limited.}
In authentication tasks, class-level information is limited, and feedback from the model is essentially non-existent, apart from the final authentication decision. Although the underlying model may be the same, the public interface through which authentication may be performed is purposefully limited to a binary classification problem. For example, the interface provided by face authentication services, such as Windows Hello, will only display a fail or success~\cite{win_hello}. This is coupled with the fact that the number of queries available to an attacker is limited, so there is an ever--vanishing window during which the attacker can gather information.
We can consider an authentication system as an oracle that outputs a binary decision based on a provided username and token pair. An attacker could theoretically extract side-channel information about a target directly from the oracle~\cite{Fredrikson2015}, but the information returned by a real-world authentication system (Yes or No) is not sufficient for proposed inversion attacks to be practical, as such attacks require either confidence information or class predictions~\cite{Papernot2016}. We avoid relying on either of these values as they are likely not available to the user in a real-world authentication system. Our attack relies entirely on the binary authentication result returned by the oracle.


\noindent
\textbf{Target information is secret and hidden.}
In previous machine learning attacks, an adversary may perturb the features of a known target, and force it to be mislabelled. These are attacks against \textit{classification}, where an attacker takes correctly-classified input features and iteratively modifies them with the goal of altering the classification result. In the face authentication domain, this takes the form of compressing a known target into a less visually perceptible version of itself so it can bypass a classifier. 
For example, images of a known target's face can be used to construct perturbations that fool
the authentication system~\cite{Sharif2016}. This is a consequence of ``fuzzy''
authentication systems, which rely on mutable properties to establish identity.
In these cases, it is attractive to reconstruct the problem into a generative
modeling problem, where the target must be reconstructed using some additional
information acquired by the adversary. However, in a system which uses physical
properties of the target to authenticate, such as physical runtime characteristics of an
end host in a system authentication task, the adversary needs direct access to the host to gather any
information.  Thus, our own attack is motivated to be completely {\em blind}. Instead of relying on any known target information, we leverage explainable AI (XAI) techniques~\cite{lime} to infer an oracle's decision boundaries. XAI techniques, such as LIME~\cite{lime}, are primarily used for generating human--understandable explanations of a machine learning model's decision. However, we find that XAI systems also grant the ability to discover an oracle's decision boundary without needing any secret information about potential victims. 

\input{images/faces}

\noindent
\textbf{Contributions.} This paper circumvents the previous limitations by formulating an attack which combines AML, query synthesis, and XAI techniques. To the best of our knowledge, this attack is the first to use XAI techniques to aid the adversary's goal of gleaning information about the decision boundary. The attack is first formalized to show that it generalizes to other authentication domains by limiting the number of queries an adversary needs to learn the model's decision boundary. The attack is then used against state-of-the-art end-host and biometric authentication systems. As illustrated in Figure~\ref{fig:faces}, the attack is effective even in a biometric authentication system where victim signatures are secret. The attack is later verified against a widely-used black-box face authentication API named Face++~\cite{facepp}. We show an adversary can circumvent defended models and achieve 93\% success rate with as few as 100 queries to the Oracle, rivaling recent attacks which relied on top-$k$ class feedback from the oracle~\cite{Sharif2016}.


\noindent
The rest of the paper proceeds as follows:
Section~\ref{sec:background} provides background information on each domain relevant to our attacks;
Section~\ref{sec:design} presents our threat model and approach;
Section~\ref{sec:eval} gives our evaluation results;
Section~\ref{sec:discuss} discusses our use of XAI and the limitations of our approach; 
Section~\ref{sec:relwork} considers related work; and
Section~\ref{sec:conc} concludes.

\input{images/approach}

%% file: images/faces.tex

\begin{figure} 
  \centering
  	\textbf{Random Pixel + LIME Perturb. Strategy}\par\medskip
    \includegraphics[width=0.40\textwidth]{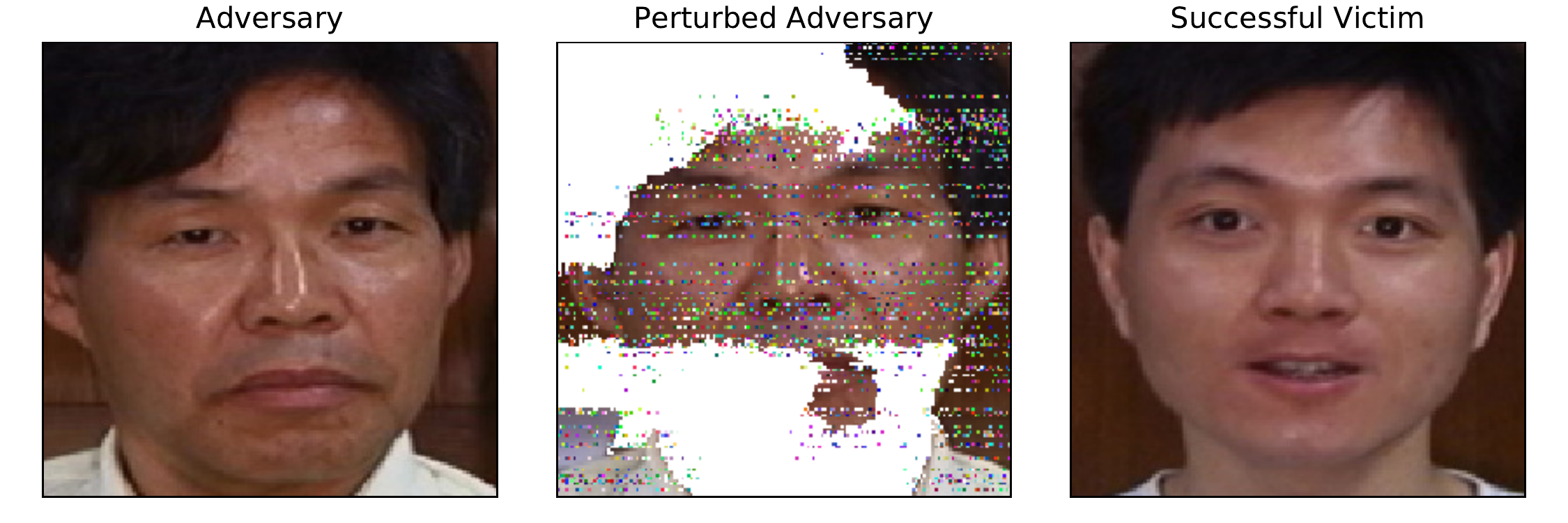}

  	\textbf{VAE Latent Space + LIME Perturb. Strategy}\par\medskip	
    \includegraphics[width=0.40\textwidth]{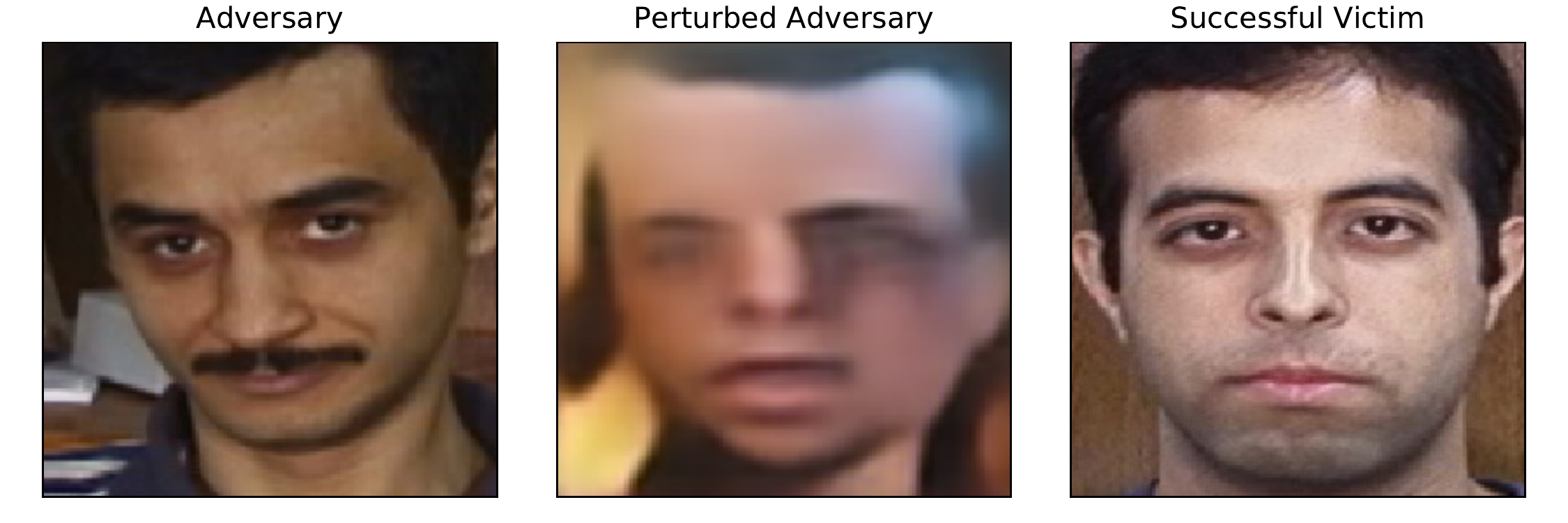}
  \caption{Examples of the two XAI-assisted perturbation techniques to fool a biometric authentication system during our proposed attack. The adversary is able to masquerade with absolutely no information of the victim, maximum success rate, and minimal queries. }


\label{fig:faces}
\end{figure}

%% file: images/approach.tex
\begin{figure*}[t!]
    \centering

    \subfloat[][Seed Dataset Creation]{\includegraphics[width=0.41\columnwidth]{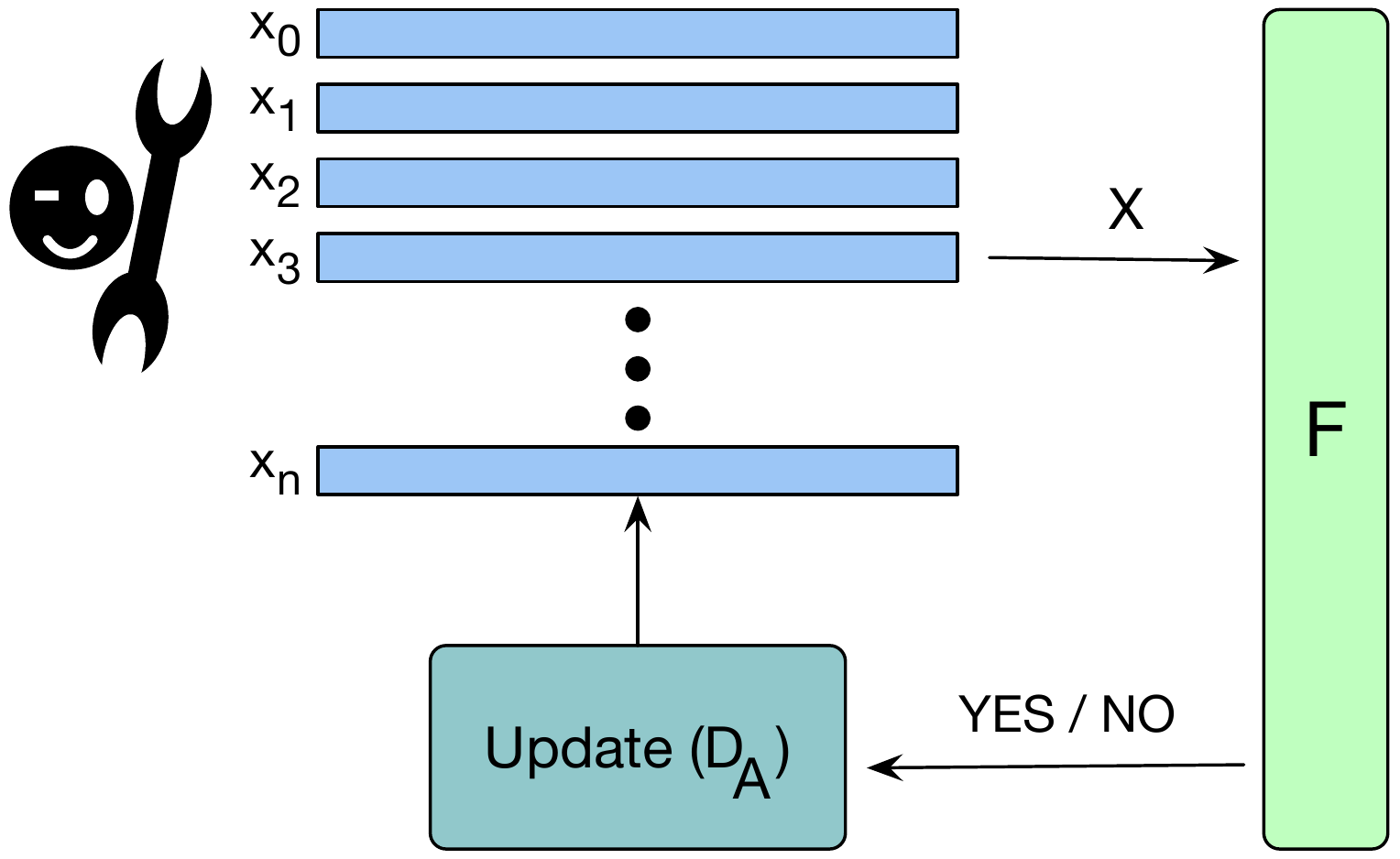}\label{fig:appr1}}
    \hspace{2cm}
    \subfloat[][Adversarial Sample Crafting]{\includegraphics[width=0.41\columnwidth]{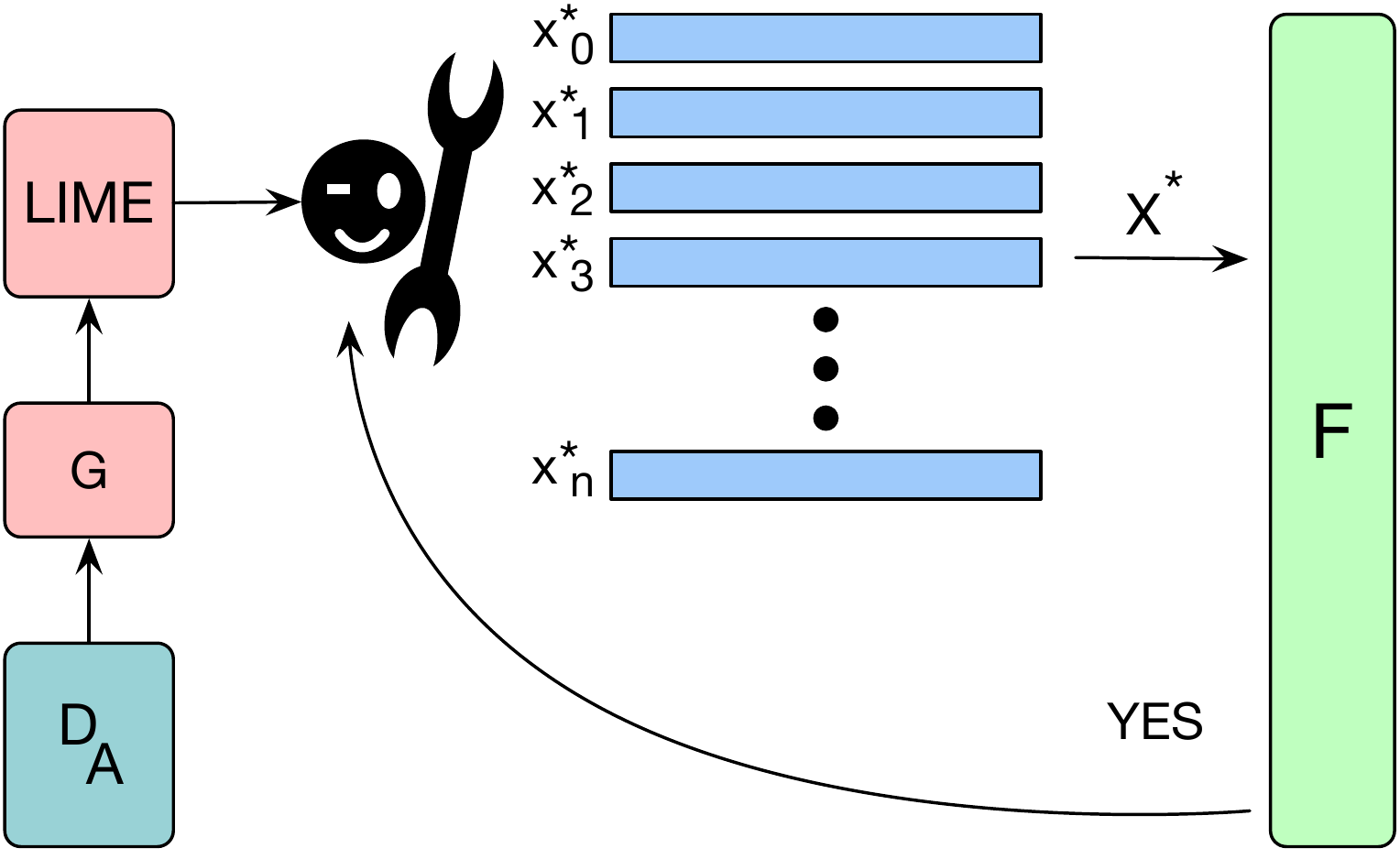}\label{fig:appr2}}
    
    \caption{Visual description of attack method. a) The adversary builds a private database of perturbed feature vectors and their corresponding labels by querying the oracle $F$ through the authentication stream $X$. Feature vectors are labelled one authentication stream at a time. b) With $D_A$ built, the adversary trains the LIME model and uses its  interpretation to synthesize a new query. This process repeats until a ``Yes'' is given. }
    \label{fig:approach}
\end{figure*}

%% file: background.tex
\section{Background}
\label{sec:background}

We focus on attacking authentication systems that use ML models to differentiate between principals. In this section, we briefly discuss each domain relevant to our attack.

\subsection{Fingerprinting for Authentication}
\label{sec:bg-fingerprinting}


Authentication is the act of verifying an identity using a set of unique credentials. Traditional credential formats include PINs, passwords, and keys. More recent authentication systems use a combination of traditional credentials and fingerprints (including biometrics). For example, recent systems are able to deduce fingerprints for smartphone users through the way they interact with the touchscreen~\cite{Jakobsson}. For computers with physical keyboards, it is possible to create a unique identity based on keystroke and mouse patterns~\cite{Mondal2017}. Each principal $u$ in a fingerprinting system is capable of generating a unique fingerprint $x_t$ at some arbitrary time $t$ after registration (hereafter known as operation {\it register}). The fingerprint may be any combination of immutable
and mutable features which are secret to all other principals in the system. The principal hereafter owns the identity $u$ and claims this identity in the future using a generated vector $z$. Given a deterministic lookup
function $F$, the objective of the fingerprinting system is to correctly evaluate the features $z_{t+i}$ claimed by identity $u$ such that $F(z_{t+i}, u) = 1$, and $F(y_{t+i}, u) = 0$ for other possible feature vectors $y$ that do not belong to $u$ at any arbitrary time step $i$.  

Authentication through fingerprints is intended to be more convenient than traditional methods, as the principal does not have to memorize or store any extra burdensome information. Rather, the principal is only responsible for generating a new sample $z$ upon request by the authentication system. 
However, the use of mutable features that act as ``fuzzy'' indicators for identity may be prone
to misclassification by the authentication system, which can be considered an estimation over the entire feature space.
Due to the nature of these fuzzy features, many samples may be necessary to make a 
decision on behalf of the principal, in which case a majority voting scheme is 
generally deployed to make the authentication decision~\cite{Bates2014}. 
This abstraction on top of the authentication decision makes the attack more 
difficult, since the adversary has less control over the model during queries.
Still, in the worst case, the mutable features could be exploited to fool the estimation of the lookup function $F$, which forms the basis for our attack.

\noindent
\paragraph{Device Fingerprinting.}
Device fingerprinting, in particular, relies on characteristics inherent in any combination of 
hardware, software, and user behavior of a particular machine.
While different in implementation, it shares some similarities with browser-based fingerprinting, a related line of work
that tracks visitors of websites based on 
browser type/version, screen resolution, language, fonts, plugins, etc. 
Device fingerprinting has been explored extensively in the literature, including by Kohno et al.~\cite{Kohno2005}, Bates et al.~\cite{Bates2014} and Radhakrishnan et al.~\cite{Radhakrishnan2015}. For example, the solution proposed by Bates et al. relies on the combined variation of USB firmware and hardware timing to generate a unique signature of the target host device, while Radhakrishnan et al. use network packet inter-arrival times. Early device fingerprinting such as the system proposed by Kohno et al.~\cite{Kohno2005} instead relied on system clock skews. Among these systems, each relied on the intuition that both soft and physical components of a device contribute a unique amount of variation that enables fingerprinting.

The key to creating a model that captures all different characteristics to a 
specific signature was ML classification techniques, such as
decision trees in the case of USB transactions, and artificial feed-forward
neural networks for network packets. However, the presence of an adversary with
malicious data samples was only partially investigated in these systems.

\noindent
\paragraph{Facial Verification.}
Facial verification, at its core, is also a form of fingerprinting.
Each face can be considered as a set of physical features which contribute
a unique amount of variation that enables fingerprinting.
Many approaches have been proposed for facial verification, including
FaceNet~\cite{facenet_tf}, which uses a deep convolutional network trained
to directly learn a mapping from face images to a compact Euclidean space,
learning directly from the face pixels. Other approaches include
DeepFace~\cite{deepface}, which uses Principal Component Analysis (PCA) to perform feature reduction, and GaussianFace~\cite{gaussianface}, which jointly learns several recognition-related tasks to learn better discriminative features.
Since a large portion of facial recognition performance relies on feature engineering, much of the literature focuses on an efficient feature representation for human faces. However, like device fingerprinting systems, facial verification systems make extensive use of feature engineering techniques that make the underlying machine learning models vulnerable to attacks.

\subsection{Adversarial Machine Learning (AML)}
\label{sec:bg-advML}

In our authentication task, the machine learning model assumes the role of lookup table as described in the previous section. The model is built as a deterministic function $F : D \rightarrow U$ mapping a training set of features $D$ to labels $U$. In our scenario, $U$ forms the labels for all principals registered with the system, and $D$ is the associated fingerprinting features that $F$ will use to estimate the distinction between principals, such that $x \in D$ for some principal's feature vector $x$. From the view of the machine learning model, the classification and authentication tasks are identical. In both cases, labels are assigned based on some criterion, and a decision is made. Many algorithms exist to create the mapping between features and labels, including Decision Trees, Neural Networks, and Support Vector Machines~\cite{Wu2008}. We investigate these models to determine their susceptibility to the attack, both before and after defenses are applied. 


In an adversarial environment, the model is susceptible to attacks that target different areas of the decision-making process~\cite{Barreno2006}. Such attacks can be either black-box, grey-box, or white-box, each denoting a successive increase in the adversary's knowledge of the underlying model. Authentication systems are considered oracles that output a binary classification based on a principal's inputs. The internal architecture and parameters of the oracle's model are not visible to the principals; thus we assume a black-box model throughout our experiments.

Our attack makes use of transfer learning, in which learned information from one task is transferred to facilitate learning of a new task (e.g., transferring the weights from a trained neural network to a new model that will be further fine--tuned). In this way, existing labeled data of related tasks or domains can be used to train new models. For image classification, general feature information on image composition is accumulated through a convolutional neural network's (CNN's) layers, where each layer is activating for a certain set of visual features. A popular technique is to take the final pooling layer of a CNN as a feature vector, as it produces a concise representation of the accumulated knowledge from every prior layer~\cite{yosinski_2014_NIPS}. This technique is powerful as vectors can be measured using some arbitrary distance function. These distances
are then used by the model to generate various outputs, as in the case of
natural language processing and face recognition models~\cite{Schroff_2015_CVPR}, as well as any
model that attempts classification over a large quantity of classes~\cite{lst_nips11}.

\subsection{Query Synthesis and XAI}
\label{sec:bg-queryAI}

We wish to minimize the amount of queries an adversary makes to the authentication system, and create queries that will correctly mis-classify the adversary as some identity $T$. Thus, we formulate our problem in the context of {\em query synthesis}~\cite{Angluin1988}. Query synthesis is a central problem in the domain of active learning, which is itself a sub-field of machine learning. The goal of active learning is to learn some concept by using as few queries, or samples, as possible. Active learning problems can arise in cases where unlabeled data is much more abundant than labeled data. For our purposes, the adversary must make minimal queries in order to learn the unknown concept of a potential victim, where the concept describes the victim's feature space. Once the adversary learns the concept, they must {\em synthesize} queries that will result in successful authentication.

In an authentication system, an adversary's only feedback is a positive or negative result. Since positive results are secret, the adversary must operate on their own data to extract a victim's concept. We leverage Explainable--AI (XAI) techniques to infer the decision boundary local to the adversary and determine the set of most influential features in the model, thus learning a concept that is useful to the adversary. Such techniques (e.g., LIME~\cite{lime}) were originally developed to present human-readable explanations of opaque model decisions, but we find they offer an intuitive interface to iteratively discover a decision boundary, and use such information to synthesize successful queries. Specifically, LIME trains a linear model using a local neighborhood of perturbations around some data point. Rather than directly performing iterative perturbations on global features as in previous attacks, we take this linear model to inform the adversary of which features to modify local to the current sample in order to move towards a successful query.


%% file: design.tex
\section{Approach}
\label{sec:design}

Given the earlier discussion, we formally describe our attack
algorithm in this section. One of the challenges is that the internals
of the ML model used by the authentication system are not available to
us. For example, if the authentication system uses a DNN that outputs
a class label corresponding to the user-id and then gives its
authentication decision (0 or 1), an attacker does not see the attack
label (he only observes whether the authentication request was
accepted or denied). We follow the formalization depicted in Figure~\ref{fig:approach}, which offers a high-level overview of the approach. 

\subsection{Formalized Threat Model \& Attack}
\label{sec:threatmodel}

\input{images/attackimpl}

\paragraph{Authentication system.} Earlier in Section~\ref{sec:bg-advML}, the model was defined as the mapping $F : D \rightarrow U$. We extend this definition to consider the dimensionality and possible decisions of the feature space, to allow for a more rigorous formalization. Let $U = \{ u_1, \cdots, u_n \}$ be the
set of current users and $F: (U \times D^k ) \rightarrow \{ 0,1 \}$
be the classifier that corresponds to the authentication system ($D$
is the domain of features with dimensionality $k$). If some user wants to authenticate as $u
\in U$, they present a vector $x \in D^k$, and the authentication
succeeds iff $F(u,x) = 1$. Formally, an authentication system (AS) is
$4$-tuple $(U,D,k,F)$. An authentication system $AS=(U,D,k,F)$ also
has an operation {\it register}, which adds a new user $u'$ to AS and
thus $U$ is updated to $U \cup \{u' \}$ and the classifier $F$ is
updated accordingly (i.e., $F$ has to accept inputs of the form
$(u',z)$, where $u'$ is the new user and $z \in D^k$ is the vector
used for authentication). The precise mechanics of the register operation
are not important to the discussion.

\paragraph{Adversary's capability, constraints and goal.} 
An adversary $\mathcal{A}$ is given black-box access to $AS$ (can
register new users and obtain answers to $F(u,y)$ for chosen $u$ and
$y$). The goal of the adversary is to impersonate some user $t \in
U$ (i.e., craft a $z$ such that $F(t,z) = 1$). There are additional
constraints that we can place on the adversary. For example, we
can limit the number of queries and the registration of new users.

Moreover, $\mathcal{A}$ can have some partial knowledge about the
vector $z$ (e.g., the make and model of $t$'s host) that will lead to
successful authentication of $t$. However, we reiterate that although $\mathcal{A}$ may have partial knowledge of $z$, it is not enough information for successful impersonation. Thus, in this paper we assume zero knowledge of $z$. $\mathcal{A}$ is permitted to have knowledge about a
subset $U_{\mathcal A} \subseteq U$ of the users (only in the case that $\mathcal{A}$ registered the users themself or might be colluding with them). For our purposes we further assume that $t \notin U_A $.

\paragraph{Concrete algorithm of the adversary.} 
Now we describe an algorithm for the adversary.  Note that at each step,
the adversary can interact with the authentication system (registering
new users or querying $F(u,\cdot)$ on vectors, which is implicit
throughout the algorithm). Essentially, the adversary is given oracle
access to the authentication system. 

\begin{itemize}
\item {\it Initial:} The adversary starts with a substitute network $G:
  D^n \rightarrow \{ 0,1 \}$ trained on {\it seed dataset} $D_A
  \subseteq D^n$, where $D_A$ is similar to a learnable coreset in the context of active learning~\cite{AL_Coreset}. We set a counter $i=0$ and set $z_0$ to be a vector,
  which is the initial guess of the attacker. The history
  $H(\mathcal{A})$ of the adversary is denoted by
  $(G,z_0,\cdots,z_i)$ and updates as the
  algorithm proceeds. In practice, this stage is implemented as the algorithm described in Algorithm~\ref{alg:quickstart}, named \textsc{QuickStart}. 

\item {\it Iterative loop:} 
\begin{itemize}
\item If $F(t,z_i) = 1$, $\mathcal{A}$ is {\sf done}. 
\item Otherwise $\mathcal{A}$ crafts $z_{i+1}$ based on its history
  $H(\mathcal{A})$.  The adversary updates its history
  $H(\mathcal{A})$ by updating $G$ (based on the fact that $F(u,z_k) = 0$ for $0 \leq k \leq i$) and increments the counter $i$. 
  Following the update, the adversary  $\mathcal{A}$ returns to the beginning of the iterative loop and continues the process until a query is successful.
\end{itemize}
\end{itemize}

Now the question is this: how does $\mathcal{A}$ craft $z_{i+1}$ from the
current history $H(\mathcal{A})$?  One option is to use existing
algorithms for crafting adversarial
examples~\cite{Papernot2015}~\cite{Papernot2016}. However, such algorithms require explicit information from the oracle about the sample's classification. For this reason, we implement our own adversarial sample crafting algorithm, called \textsc{LIME--Sampler}, which is discussed in detail in the next section. Start with $z_i$ and
the substitute network $G$ and craft an example $w$ such that $G(w) =
1$ and set $z_{i+1} = w$. $G(w) = 1$ implies that $w$ is crafted so it eventually inhabits the feature space nearby some fingerprint $x$ for target identity $t$ (with $F(t, x) = 1$). The algorithms for constructing adversarial
examples can perform better if they rank features
that contribute most to the decision.  For the purpose of describing
the influence of features on the decision of the substitute model $G$,
we use techniques from the  XAI literature. Specifically, our algorithm
leverages an XAI system, LIME~\cite{lime} to rank
the features that correspond to $G$'s decision on $w$.

\subsection{Constructing the Seed Dataset}
\label{sec:build_sd}

As described in the algorithm above, the adversary must first build a seed
dataset $D_A$ which is used to train the substitute model $G$. There are
two challenges in constructing $D_A$. First, $D_A$ should have enough
positive and negative examples such that new information is deduced, rather than memorized, about the cause for a ``Yes'' or a ``No''. Second, as we alluded to before, we can only rely on ``Yes'' and ``No'' decisions from the authentication system, rather than class--level information. The idea of discovering the smallest fundamental set from which to learn a feature space is known as coreset construction in active learning, and thus we frame our discussion in a similar context~\cite{AL_Coreset}.

The key factor in an adversary's ability to infer the decision
boundary is the quality of the adversarial sample crafting
algorithm. To illustrate this, we start with a simple case following the notation of the previous section. Let $D$ be some sample crafting algorithm which takes as input the adversary's candidate feature vector, and outputs a highly distorted version of this vector. For the purpose of discussion, the exact heuristic to measure distortion is not important, we simply assume that the distortions produced by $D$ mostly result in outliers within the feature space learned by $F$. The adversary may construct some feature vector $z_{i} = D(z_{i-1})$ for $i > 0$, forcing $F$ to associate the adversary with the closest principal $t$ whose feature vector resembles $z_{i} = D(z_{i-1})$, (that is, $F(z_{i}, t) = 1$). Inuitively, $t$ is an outlier who unfortunately is closest in the feature space to the distortions created by $D$. However, we assume for any model that the feature space is sufficiently normalized to minimize the number of outliers. If we limit the adversary's
knowledge to only a subset of usernames $U_A \subset U$, and $t \notin U_A$, in this case the adversary will not have any
chance of success. We can further deduce that if the amount of outliers in the feature space is minimized, and the adversary's knowledge requirement is also minimized, the probability of success quickly diminishes. In the opposite case, where $z_{i}$ closely resembles the adversary's own feature vector, they will always be classified as themselves. It follows that if
the adversary crafts ``No'' samples, and their distortions are too
similar to ``Yes'' samples, it will be difficult to model their
distinction. 

We must thus discover some learnable subspaces for ``Yes'' and ``No'' in the total feature space. This motivates the creation of an iterative sample
crafting algorithm, \textsc{QuickStart}, which can strike a balance between the two extreme cases discussed above.  One of the challenges in
designing \textsc{QuickStart} is that we must rely only on the
oracle's authentication decision to increase or decrease distortion.
\textsc{QuickStart} is described in Algorithm~\ref{alg:quickstart}.

\input{algorithm/quickstart}

Unlike previous algorithms, \textsc{QuickStart} relies only on a
binary decision from the oracle it is querying. The primary goal is to
build a dataset of \textsc{YES} decisions denoted by
$D_{\mathcal{A}_{\mbox{yes}}}$ and a dataset of \textsc{NO} decisions denoted
by $D_{\mathcal{A}_{\mbox{no}}}$, which combine to form $D_\mathcal{A}$. We constrain the algorithm so that each set is balanced, such that $|D_{\mathcal{A}_{\mbox{yes}}}| = |D_{\mathcal{A}_{\mbox{no}}}|$ (for the purpose of better performance when training $G$ later). The
attacker starts with a set of benign samples they own.

The {\it first loop} of the algorithm is responsible for dividing
the samples from the set $A_{GT}$ (the set of initial samples) into the set $D_{\mathcal{A}_{\mbox{yes}}}$.

The {\it second loop} performs the process of discovering the set of
$\textsc{NO}$ samples (if we had enough negative samples from the
first loop, the second loop does not execute).  This loop relies on a
procedure $\textsc{Perturb}(x, \Upsilon)$, which perturbs the vector
$x$ according to the parameter $\Upsilon$ (higher $\Upsilon$ means
that the perturbation has higher variance).  The lower and upper
bounds of the exponentially increasing distortion form a sliding
window that can be easily controlled by the lower bound $\beta$ and
upper bound $\alpha$. For example, the adversary may observe their own
features and use a fraction of the lower and upper bounds
available. The strength of the trained model $S$ is ultimately
determined by the number of training samples harvested during
\textsc{QuickStart}.  Although a larger number of samples is
preferable depending on the architecture of $G$, a primary goal of $\mathcal{A}$ is to also minimize
queries; thus, the size of sets $D_{\mathcal{A}_{\mbox{yes}}}$ and
$D_{\mathcal{A}_{\mbox{no}}}$ is bounded by a tunable parameter $\sigma$.

\subsection{Crafting Attack Samples}
\label{sec:build_as}

Once $\mathcal{A}$ has the dataset $D_\mathcal{A}$, it must train a
substitute model $G$ as shown in Figure~\ref{fig:attackimpl}. This step is
inspired by black-box attacks which exploit transferability properties
of oracle decision boundaries~\cite{Papernot2016}. Rather than use
explicit class labels, we rely on the \textsc{YES} and \textsc{NO}
decisions made by the oracle with respect to perturbed data to create
a model of the decision boundary. The goal of the adversary is to pick
a model architecture that does not require a large $D_\mathcal{A}$,
but can make a reasonable assumption about future adversarial samples
$z_{i+1}$. In practice, the primary restriction on the attack's
effectiveness is the selection of $G$ and size of $D_\mathcal{A}$. For
example, with a shallow neural network, the accuracy of $G$ ranges
from 30\%-80\% depending on the sample type, size of $D_\mathcal{A}$,
and quality of samples returned by \textsc{QuickStart}. Although we
optimize our network architecture for test-set performance, the
limited size of $D_\mathcal{A}$ can limit the effectiveness of
$\mathcal{A}$ based on its similarity to other principals. However,
once $G$ is trained, $\mathcal{A}$ can easily transform this model
into guided perturbations using an XAI system such as
LIME. In 1-dimensional feature modalities, LIME can be used to
describe specific feature values that will contribute to either oracle
decision. Specifically, LIME outputs an interval the feature is expected to occupy for its associated weight on the decision's class. Using this information, $\mathcal{A}$ crafts attack samples
$z_{i+1}$ and attempts to masquerade as some other principal in the
system. The algorithm to perform this process is outlined in Algorithm~\ref{alg:limesampler}. 

The main goal is to synthesize a new feature vector $x* \in \mathbb{R}^k$ by modifying the $r$ most relevant features in $x \in \mathbb{R}^k$ which contributed to the \textsc{NO} class. It follows that $r$ can be tuned to adjust the distortion of $x*$, such that $0 < r \leq k$ for the feature space $D^k$. For the sake of the algorithm, we define an explanation returned from \textit{LIME--Query} as a vector of 3--tuples containing the index of a feature, the weight of the feature on a \textsc{NO} decision, and the interval it is expected to occupy. Thus we have an explanation $E_x = \{ (0, {w}_0, [c_0, d_0]), ..., (k, {w}_k, [c_k, d_k]) \}$ for feature index $i \in \{ 0, ..., k \}$, feature weight $w_i \in \mathbb{R}$, interval lower bound $c_i \in \mathbb{R}$, and its respective upper bound $d_i \in \mathbb{R}$. The top $r$ features and their intervals are found by simply sorting $E_x$ on the values of $w_i$, and taking the top $r$ 3--tuples. To arrive at the new value for feature ${x*}_i$ with $i \in \{0, ..., r\}$, we sample uniformly from a feature's expected interval, such that ${x*}_i \sim U(c_i, d_i)$. We further define a set of principals $V$ for which $\mathcal{A}$ knows the usernames, and further assume that $v \in V$ with $V \subset U$. Thus, Algorithm~\ref{alg:limesampler} allows us to affect the most relevant features in $x$ that allow $\mathcal{A}$ to masquerade as some other principal, and stop as soon as a successful query is encountered.

\input{algorithm/limesampler}

%% file: images/attackimpl.tex
\begin{figure*}[t!]
    \centering
    \includegraphics[height=1.4in]{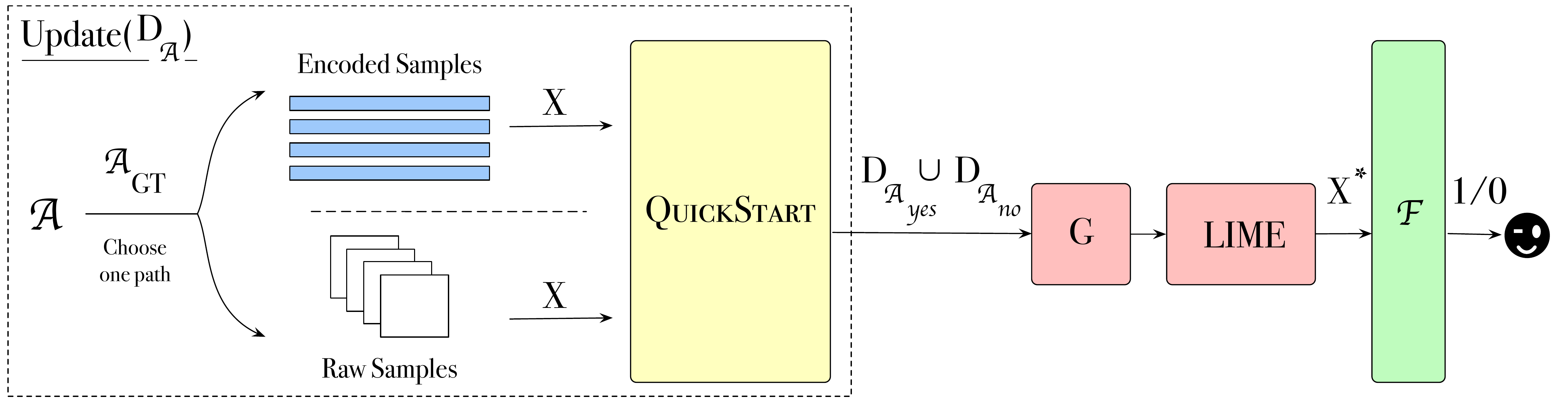}
    \caption{Workflow for building attack samples.To update $D_A$, the adversary applies ground truth as either a VAE-encoded vector, or raw images. The output from this step goes into QuickStart. $D_A$ is
passed along through $S$ and LIME to issue $X^*$ to oracle $F$.}
    \label{fig:attackimpl}
\end{figure*}

%% file: algorithm/quickstart.tex

\begin{algorithm}
\caption{\textsc{QuickStart}, seed dataset builder.\\
\textbf{Input: } $A_{GT}$, a set of initial samples owned by $\mathcal{A}$. $\beta$,
exponential distortion lower bound. $\alpha$, exponential distortion
upper bound. $\sigma$, the desired number of samples for each class (negative
and positive), $O$, the victim Oracle, and subroutine \textit{query}, a generic interface available to \textit{A} for querying $O$.}
\label{alg:quickstart}


\begin{algorithmic}

\State $D_{A_{yes}}, D_{A_{no}}$ $\gets$ $\varnothing$
\State $R \gets $ \textsc{YES} 

\While{$\mid D_{A_{yes}} \mid <  \sigma_l$}
	\State $x \gets _RA_{GT}$ $\rhd$ (\text{random sample from } $A_{GT}$) 
        \State $R \gets \textit{query}(O, A, x)$ 
        \State \textbf{if }{$R = \textsc{NO}$}: \textit{add}$(D_{A_{no}},x)$
	\State \textbf{else }: \textit{add}$(D_{A_{yes}},x)$
\EndWhile

\State $R \gets $ \textsc{YES}, $\Upsilon \gets 0$

\While{$\mid D_{A_{no}} \mid < \sigma_l$}
	\State $x \gets _RA_{GT}$ 
	\State $\rhd$ (\text{Loop until we meet sufficient distortion})
	\While{$R =$ \text{ YES}} 
		\State \textit{increment}($\Upsilon$)	
		\State $x* \gets \textsc{Perturb}(x, \Upsilon)$
		\State $R \gets \textit{query}(O, A, x*)$	
	\EndWhile

	\State $x* \gets \textsc{Perturb}(x, \Upsilon)$
	\State $R \gets \textit{query}(O, A, x*)$	
	\State \textbf{if }{$R = \textsc{NO}$}: $\textit{append}(D_{A_{no}},x*)$


\EndWhile
\State return $D_{A_{no}}, D_{A_{yes}}$ \\

\end{algorithmic}

\smallskip









\end{algorithm}

%% file: algorithm/limesampler.tex

\begin{algorithm}
\caption{\textsc{LIME--Sampler}, adversarial sample crafting algorithm.\\
\textbf{Input: }
$D_A$, the seed dataset created with \textsc{QuickStart}, $G$, an arbitrary model that can be trained on $D_A$, \textit{train}, the associated training function for $G$, \textit{LIME--Query}, the subroutine provided by LIME to query given a trained model and candidate sample, and subroutine \textit{query}, a generic interface available to \textit{A} for querying $O$.}
\label{alg:limesampler}

\smallskip

\begin{algorithmic}

\State $G \gets $ \textit{train}($G$, $D_A$)
\State $R \gets $ \text{NO}

\While{$R = $\text{NO}}
	\State $x \gets _RA_{GT} $ (\text{random sample from } $A_{GT}$)
	\State $\rhd$ (\text{Get explanation tuples $E_x = \{ (0, {w}_0, [c_0, d_0]), ..., (k, {w}_k, [c_k, d_k]) \}$ tuples for \text{NO}})
	\State $E_x \gets $ \textit{LIME--Query}($x$, $G$, \text{YES}) 
	\State $\rhd$ (\text{Sort over feature weights $w_i$ and take top $r$ tuples})
	\State $E_x \gets $ \textit{sort($E_x$)$[:r]$}
	\For{$i$, ${w}_i, [c_i, d_i] \in E_x$ }
		\State $\rhd$ (\text{Update feature $x_i$ with uniform sample over expected interval.}) 
		\State $x_i \gets U(c_i, d_i)$ 
	\EndFor
	\State $v \gets _RV $ (\text{random sample from } $V$)
	\State $\rhd$ (\text{Now attempt to authenticate as principal $v$})
	\State $R \gets \textit{query}(O, v, x)$	

\EndWhile










\end{algorithmic}

\smallskip

\end{algorithm}

%% file: eval.tex
\section{Evaluation}
\label{sec:eval}

Our attack is targeted against both host and biometric authentication systems. Our experiments are motivated by the following questions:

\begin{enumerate}
	\item Are classification models suitable for use in authentication systems, even with a relatively low number of principals and private data?
	\item Are perturbations during the attack, and their effects, understandable by humans?
	\item Do defenses such as adversarial sample injection afford classification models robustness against masquerade attacks?
\end{enumerate}

We answer these questions by attacking models and data that have been vetted by the research community. For clarity, we describe the implementation, mechanisms, and data used for each target system. 

\subsection{Experimental Highlights}
\label{sec:exp-highlight}

From our experiments, we observe that the attack is effective across the tested domains and machine learning classifiers. In particular:

\begin{enumerate}
	\item We find that classification models, even when defended, are vulnerable to our attack, giving the adversary up to 93\% success rate depending on the model architecture. The attack is equally effective against a commercial facial recognition system.
	\item Perturbations can be understood by the adversary in both domains. For host authentication, the adversary simply notes the USB enumeration timings which produced success. In biometric authentication, the adversary can use LIME's graphical plotting capabilities to reveal portions of the image that contributed to successful masquerade. 
	\item Adversarial training is effective for certain models, and greatly reduces the adversary's probability for success. However, future adversarial inputs could be easily modified to evade such training.
\end{enumerate}

\subsection{Data and Implementations}
\label{sec:eval-data}

\noindent
\paragraph{Host authentication.} We evaluate against a host authentication system based on USB enumeration timings of a computer under test~\cite{Bates2014}. This system uses a Random Forest as the underlying machine learning model, trained in a \textit{target} vs. \textit{outlier} fashion, with a new model created for every principal registered to the system. Outlier classes are balanced with respect to every possible principal in the system. Over-sampling is performed with replacement until the number of target samples matches the number of outlier samples. In this way, there are equal numbers of class data samples registered in the machine learning model. The implementation and data was obtained from the authors of the original system~\cite{Bates2014}. The trace data for this system consists of identical machines across nine users. This data also had scripts to generate the accompanying datasets for each principal's model. As described in~\cite{Bates2014}, certain hosts must be filtered from the dataset to remove false positives. We note that two of the nine hosts were removed from the set of principals after applying this constraint. Due to the limited sample size of this dataset, we focus much of our experiments on biometric authentication. Nevertheless, we believe this system's scope is realistic, and helps us generalize the attack to similar host authentication schemes~\cite{Kohno2005,Radhakrishnan2015}.

\noindent
\paragraph{Biometric authentication.} We target a facial authentication system based on a state-of-the-art facial recognition model named Facenet~\cite{Schroff_2015_CVPR}. The implementation is taken from a popular open-source repository on GitHub~\cite{facenet_tf}. We use only pre-trained models provided by the repository's maintainer. The Facenet embedding model was trained on a subset of the Microsoft Celeb-1M image dataset~\cite{1M_celeb},
which was released for the MSR Image Recognition Challenge at ACM Multimedia 2016.
Since Facenet may act as an embedding model underneath an authentication model, we train a Support Vector Machine classifier as described by the repository's documentation to perform authentication (denoted Facenet-SVM). For completeness, we also measure the attack against two additional models: Random Forest (Facenet-RF) and Neural Network (Facenet-NN), to understand the attack's scope in systems which have different architectures. These models were selected as they demonstrated the best classification accuracy when combined with Facenet. After training, none of the authentication systems exhibited false positives on hold-out evaluation data.

Training and evaluation data is taken from the Georgia Tech Face Database~\cite{gtfb} and contains faces of 50 people, with 15 images per person. 
The provided faces are composed of frontal shots that vary in facial expressions and lighting conditions, which we deem realistic for a real-world face authentication system. For consistency, this image data is pre-processed with Multi-task CNN~\cite{mtcnn}, the same face alignment model used to pre-process all Facenet-related training data. We make no other assumptions about the model, and argue that the adversary would be free to use any face-cropping tool available to have consistency in their attack. 

Due to the large feature space of face images, we propose two different paradigms for perturbing attack images, which are demonstrated in Figure~\ref{fig:faces}:

\begin{enumerate}
	\item \textbf{Random Pixel.} This attack implements \textsc{Perturb} from Figure~\ref{alg:quickstart} as the adversary setting a subset of pixels to random subpixel values according to $\Upsilon$, inspired by recent state-of-the-art attacks~\cite{Papernot2015, Papernot2016}. In this attack, the entire image constitutes a single sample fed into $G$. 

	\item \textbf{VAE Latent Space.} This attack makes use of the Facenet Variational Autoencoder (VAE) to create natural and realistic attack images. Similar to hill-climbing attacks which target feature dimensionality reduction~\cite{hillclimbingICB, hillclimbing09}, the adversary implements \textsc{Perturb} to modify the low-dimensional latent vector created after encoding their face with the VAE, similar to a natural image attack proposed by Zhao et al.~\cite{zhao2018generating}. This means the adversary is able to create natural and believable face samples with feature distortion $\Upsilon$. Perturbed latent vectors are used as training data for $G$, and are decoded to query the authentication system. 

\end{enumerate}

Thus, we offer two different techniques for perturbing adversary data before feeding it into LIME. We take the GitHub repository's version of the Facenet VAE along with the provided weights trained on the CASIA-WebFace dataset~\cite{casia_webface}, a large-scale dataset containing almost 500k faces of over 10k people. We clarify that every pre-trained model used in the experiments was trained on a different and unique dataset. That is, we assume the adversary has no access to the oracle's training data, other than what the adversary has registered. 

The use of a Facenet-derived generative model may partially violate the black-box assumption. To verify our attack, we run experiments against a face recognition API named Face++~\cite{facepp}. To clarify, Face++ does not have any relation to Facenet, apart from their similar naming. We build a small oracle abstraction around the Face++ API to return only a \textsc{YES} or \textsc{NO} to the adversary. A configurable parameter of Face++ is the threshold at which it returns a positive result. If the provided face aligns with an identity in the top $k$ matched identities, it will return a \textsc{YES}. Unlike previous work~\cite{Sharif2016}, we configure the Face++ API's top-$k$ response to $k=1$ and only count a query as successful if the top result returned by Face++ matches the identity claimed by the adversary. We believe this scenario is the most realistic, as it targets the most secure setting of $k$ possible within the system, and is simultaneously the worst-case scenario for the adversary.

\subsection{Metrics}
\label{sec:metrics}

We define a successful attack as one in which the adversary is able to masquerade as some other principal on the system, that is, {\em reach masquerade}, and denote probability of masquerade as $P(M)$. 
A principal's login credentials can be interpreted as some pairing of username $u$ and a unique secret token $d$ initially known only by the owner of $u$. It follows that if $u$ is known, an adversary may attempt to generate $d$ for an observable feature space $D^k$ where $d \in D^k$, defined prior in Section~\ref{sec:threatmodel}. As the adversary's knowledge of the usernames in the population $U$ increases, so does success probability $P(M)$. Let $V$ denote the set of known \textit{usernames} (not secret tokens) by the adversary, such that $V \subset U$. With complete knowledge of all usernames ($V$ = $U$), the adversary has success probability $P(M) = 1$, assuming the adversary has infinite queries available and may try every combination of features available. It follows that for $V = \emptyset$, $P(M) = 0$. Due to practical computation restrictions, this assumption is not very useful for the adversary. However, we show that $P(M)$ may be maximized as long as $V \neq \emptyset$.

With some arbitrary amount of knowledge of the system's registered usernames, $V$, and a realistic bound on feature combinations, the adversary desires to increase probability $P(M)$ and minimize $||V||$. The most straightforward way to accomplish this is to artificially increase the distribution of false positives in the system, such that $P(FP_O) >> P(FP_R)$ for the optimized approach $O$, and random approach $R$. XAI systems such as LIME enhance the adversary's knowledge of the oracle's decision boundary, allowing the adversary to increase $P(M)$, thus minimizing the number of queries needed to find a false positive. Thus, for every experiment we measure the number of queries at which the adversary achieves masquerade, and the adversary's authentication decision for every victim $u \in U$. $P(M)$ is calculated as the number of principals who could achieve masquerade, divided by the total number of principals in the experiment. Thus, if there is a set of principals in the experiment $U$, and the set of principals who achieved masquerade $U_M$, we have $U_M \subseteq U$ and $P(M)_{AS} = \frac{||U_M||}{||U||}$ for an authentication system $AS$.

\subsection{Experiments}
\label{sec:eval-design}

We run experiments to measure the success rate and number of queries for different versions of the approach across different domains. Each experiment measures the authentication system's decision and number of queries from the scope of both adversary and the principal (i.e., the victim):

\begin{enumerate}
	\item \textit{Baseline.} The baseline experiment establishes metrics when the adversary uses their own unmodified data against the victims.
	\item \textit{Naive Replay (host authentication).} The adversary exploits assumptions that host authentication systems make about variance in the principal's authentication stream~\cite{Bates2014,Radhakrishnan2015,Dalai2017}. The adversary replays one sample of their own data enough times to form the authentication stream, and forms the most effective, yet naive comparison to our attacks. 
	\item \textit{Naive Random (biometric authentication).} The adversary samples $x \sim \mathcal{N}(\mu, \sigma^2)$ for every feature $x \in X$ in the attack sample. This experiment is the same in both Random Image and VAE Latent Space attacks, such that a potential adversary naively samples random subpixels to create an image. 
	\item \textit{QuickStart. } The QuickStart experiment measures the metrics in Section~\ref{sec:metrics}, $P(M)$ and number of queries, when only the \textsc{QuickStart} algorithm is used.
	\item \textit{QuickStart+LIME.} The QuickStart+LIME experiment measures the same metrics as the QuickStart experiment but when \textsc{QuickStart} is combined with the LIME XAI system to enhance perturbations. 
\end{enumerate}

\input{images/usb_mats}

\input{tables/usb_que_tab}

\input{images/usb_dist}

\subsection{Host Authentication}
\label{sec:usb-auth}

In Figure~\ref{fig:usb_mats}, the authentication success of each adversary and victim pair is displayed for each experiment. The baseline displays an expected matrix for a proper authentication system, with no false positives and no false negatives. The replay experiment shows the effect of an adversary performing the naive replay attack described in Section~\ref{sec:eval-data}, such that an adversary simply replays one of their samples until they form a valid input stream. We note that in this attack, a potential adversary has $P(M)=100\%$, but the attack holds little utility because they are still identifiable by the sample passed to the system. The oracle could also easily enforce a minimum level of variance across incoming input streams before performing authentication. Thus, this type of attack is easily mitigated.

With the QuickStart attack, adversaries yield successful \textsc{YES} queries and are no longer identified as themselves. Several instances of previously unsuccessful adversary-victim pairs now yield a positive result in favor of the adversary. In total, five of the seven principals in the system were vulnerable to potential adversaries, with one adversary gaining access to two separate identities. The QuickStart+LIME attack yields a lower success rate than QuickStart alone, as shown in Table~\ref{tbl:usbqueries}. The number of queries for each are within standard deviation of each other ($112.25 \pm 41.67$ and $133.25 \pm 34.64$, respectively). We believe the limited feature space of this dataset are limiting factors for LIME's effectiveness. 

To further investigate why the QuickStart attack is effective, a victim's real trace and a successful imposter trace are plotted in Figure~\ref{fig:usb_dist} to examine their difference in feature distribution. The distribution is binned such that each bin corresponds to the nearest time of an enumeration feature, in milliseconds. Previous work tested robustness of authentication systems by attempting to replay samples with specific feature distributions that mimicked the target~\cite{Radhakrishnan2015}. However, Figure~\ref{fig:usb_dist} reveals that an exact replica of the feature distribution is not needed. The imposter sample has a larger frequency of enumeration timings in the zero millisecond and two millisecond bins, and was still misclassified as the victim. The attacker must simply exploit one {\em blind spot} in the decision boundary to launch an attack. 


\subsection{Biometric Authentication}
\label{sec:face-auth}

The performance of the attack varies widely when targeted against Facenet-based classifiers. We first examine each model separately without defenses (row one of each coverage map in Figures~\ref{fig:svm_mats},~\ref{fig:forest_mats}, and~\ref{fig:nn_mats}, respectively). For all experiments, we take the adversary's knowledge as 30\% of the total registered usernames in the system for ease of presentation (thus $||U||=50$ and $||V||=15$ for an attack against the entire Georgia Tech Face Database~\cite{gtfb}). We experimentally observed the same general trends for our attack at lower and higher knowledge levels, and leave the optimization of knowledge level for future work. As discussed in Section~\ref{sec:metrics}, $P(M)$ is tied to the size of adversary's knowledge set $V$. Intuitively, the attacker desires the highest possible coverage of false positives across principals, as this maximizes $P(M)$ when a smaller knowledge set $V$ is taken. For example, if an attack yields only a single false positive for some set $U'$ of identities, and the adversary takes a smaller subset $U'' \subset U'$, such that $||U''|| << ||U'||$, it is unlikely that the subset will yield a successful attack. 

\subsubsection{Support Vector Machine (Facenet-SVM)}

\input{images/face_mats}

The attack coverage is shown in Figure~\ref{fig:svm_mats} for the random pixel and VAE attacks. The baseline map for both versions exhibits the expected classifications for an authentication system, confirming the model's accuracy at test time. The model begins to misclassify when random inputs are used, yielding a high success rate, as observed in Table~\ref{tbl:avgqueries-all}. However, upon closer inspection of Figure~\ref{fig:svm_mats}, every adversary is misclassified as the same victim. In this case, the distribution of false positives is not consistent and gives a low chance for a successful attack if a new subset of victims was chosen. Although calculated $P(M)=87\%$ for QuickStart is lower, false positives are more evenly distributed in the system. This distribution further increases when LIME is used, yielding both high attack coverage and high $P(M)$ of 80\%. The VAE latent space attack is very effective and improves upon the random pixel attack. In the VAE attack, QuickStart and QuickStart+LIME perform at the same $P(M)$ of 87\%.

\subsubsection{Random Forest (Facenet-RF)}

\input{images/forest_mats}

The attack coverage against the RF classifier is shown in Figure~\ref{fig:forest_mats}, and exhibits a higher vulnerability in general than SVM. The random attack exhibits the same behavior where adversaries are often classified as the same victim. QuickStart dramatically improves the attack coverage in both the Random Pixel and VAE attacks, with up to 100\% success rate as seen in Table~\ref{tbl:avgqueries-all}. Perceptually, QuickStart+LIME has higher coverage for the random pixel attack than for the VAE attack.

\subsubsection{Neural Network (Facenet-NN)}

As before, the attack is successful against the NN classifier, as shown in Figure~\ref{fig:nn_mats}. Although the random attack yields high success rate, victims are focused on a single identity. The QuickStart attack improves the coverage in both Random Pixel and VAE attacks, yielding $P(M)$ of up 87\% in both attack types according to Table~\ref{tbl:avgqueries-all}. The VAE Latent Space does not benefit from LIME as much as the Random Pixel attack, where $P(M)$ increases by $27\%$ with LIME.

\input{images/nn_mats}

\input{images/all_tbl}

\subsection{Defenses}
\label{sec:defenses}

We consider our attack against biometric authentication when defenses derived from the literature~\cite{Papernot2016} are added to the oracle. The selection of defenses is not meant to be exhaustive, but rather examine whether our attack can be mitigated by model-agnostic implementations. 

\subsubsection{Injecting Random Noise (Random Defense)}
\label{sec:rand_noise}

Rather than initializing the oracle on the normal training data, we also inject a new class named \textit{other} to the training set, which consists of images with randomly generated pixel values. The number of images in this class is chosen to match the maximum number of training set images for each principal in the dataset. Thus, when an adversary attempts to launch an attack using such images, they are identified as \textit{other} and a \textsc{NO} is returned. This defense corresponds to the second row of each coverage map in Figures~\ref{fig:svm_mats},~\ref{fig:forest_mats}, and~\ref{fig:nn_mats}. The defense completely blocks the naive random attack for every model tested, while also slightly mitigating the effectiveness of the Random Pixel attack. In this scenario, the benefit of LIME is most clear as it offers the best coverage against an oracle with this defense. Since this defense does not target adversarial images from a generative model, it is not effective at blocking the VAE Latent Space attack. 

\subsubsection{Injecting Random Noise + Fakes (All Defenses)}
\label{sec:rand_fakes}

In addition to training on noisy data, the oracle is also given a class named \textit{fake} that consists of adversarial samples created by a generative algorithm. This defense must be implemented carefully, as the oracle should still be robust enough to properly classify principals that are partially occluded or blurry. Thus, we devise a defense that will demonstrate whether QuickStart and QuickStart+LIME are producing useful, non-random perturbations to the latent space. The defense is initialized by first using the Facenet-VAE model to encode one of every principal's images into the latent space. We note that in this scenario, the oracle has the advantage of owning the generative model used by the adversary. If this were not true, the defense would still be valid, so long as the chosen generative model can create fake, yet realistic images of the principals. The oracle takes the maximum and minimum values of the selection's encodings to create a lower and upper bound of latent variables for each principal. These bounds are used to randomly sample new latent space vectors, which are decoded to form an adversarial example. Thus, we will observe if perturbations induced by QuickStart and QuickStart+LIME are either meaningful deductions from the adversary's original data, or simply random samples in the latent space. This defense corresponds to the third row of each coverage map in Figures~\ref{fig:svm_mats},~\ref{fig:forest_mats},~\ref{fig:nn_mats}. QuickStart and QuickStart+LIME circumvent defenses for all models, and reduce the adversary's $P(M)$ by up to 67\% in the case of the SVM classifier, which saw a performance drop once LIME was introduced. Overall, LIME tends to improve the attacker's odds when the feature space is larger. This is expected as LIME is largely data-driven in its implementation, and suffers when a smaller amount of information is available to derive the decision boundary.

\subsection{Black-box Attack (Face++ API)}

In general, our results suggest that QuickStart is able to circumvent defenses which inject adversarial training data into the oracle. In terms of query count, QuickStart and QuickStart+LIME are within standard deviation of each other. As such, the main benefit of using LIME is a better attack surface, depending on what classification model the oracle uses. We believe the QuickStart+LIME in the VAE Latent Space configuration gives an adversary the highest chance of launching an attack, as in the worst case it will perform slightly worse if the oracle uses an SVM classifier, but better if the model is Random Forest or Neural Network. 

To verify our results, we target the Face++ face recognition API as described in Section~\ref{sec:eval-data}, providing it the entire Georgia Tech Face Database~\cite{gtfb} as training data (as such, $||U|| = 50$). We use the VAE Latent Space attack with both QuickStart and QuickStart+LIME, to compare their effectiveness on a black-box oracle, assuming the same adversary knowledge of $||V||=15$ from previous experiments. The coverage map for this experiment is shown in Figure~\ref{fig:facepp_mat}, and queries from each experiment in Table~\ref{tbl:ppqueries}. We note that the naive random attack is not successful, suggesting that Face++ protects against images with randomly generated pixels. The QuickStart and QuickStart+LIME attacks are both successful, with QuickStart+LIME yielding better attack coverage than QuickStart alone, which mimics the earlier experiments against Facenet-NN.

%% file: images/usb_mats.tex
\begin{figure}[t!]
  \centering
    \text{USB Enumeration Perturb. Attack Coverage}\par\medskip
    \includegraphics[width=0.49\textwidth]{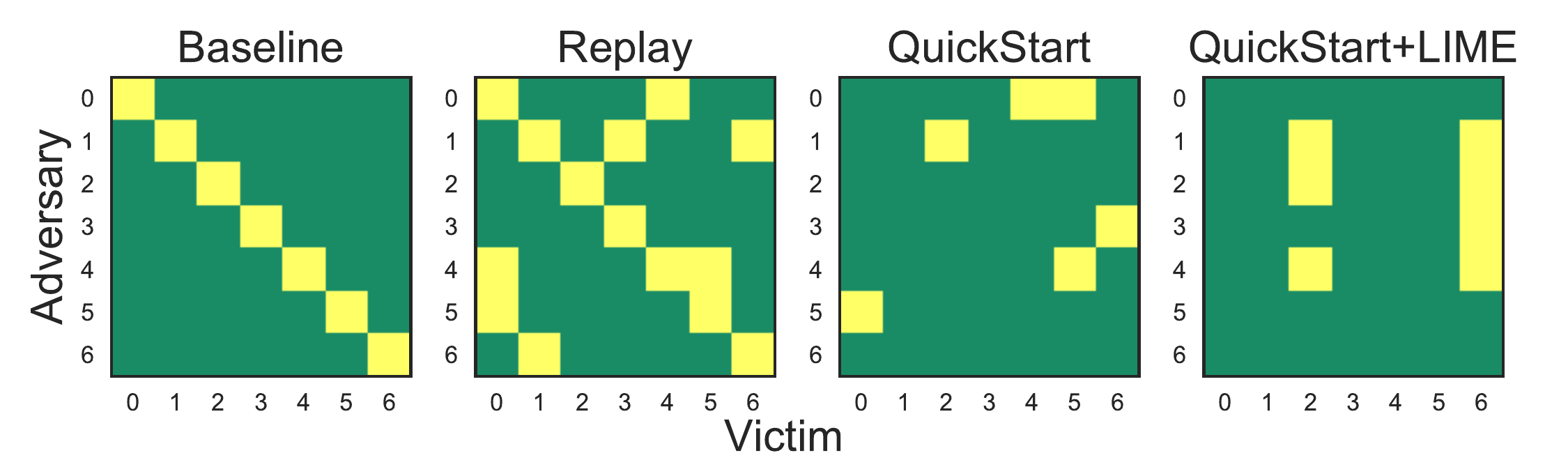}

  \caption{Attack coverage map based on authentication success for adversary--victim pairs. An adversary knows all seven usernames in the system during the attack.}

\label{fig:usb_mats}
\end{figure}

%% file: tables/usb_que_tab.tex
\begin{table}[t!]
\setlength\tabcolsep{3pt}
\footnotesize
\centering
\begin{tabular}{|c||c|}
\hline
 Replay   & 57.25 $\pm$ 65.99 \\
& $P(M)=71$\%   \\\hline
 QuickStart & 112.25 $\pm$ 41.67 \\
& $P(M)=71$\%  \\\hline
 QuickStart     & 133.25 $\pm$ 34.64 \\
+LIME& $P(M)=57$\%  \\\hline
\end{tabular}
\par\medskip

\caption{The average number of adversarial queries before success in each of three experiments,
testing against USB enumeration at 100\% knowledge.
$P(M)$ is the calculated probability based on attack coverage for each experiment.}
\label{tbl:usbqueries}
\end{table}

%% file: images/usb_dist.tex
\begin{figure}[t!]
  \centering
    \includegraphics[width=0.40\textwidth]{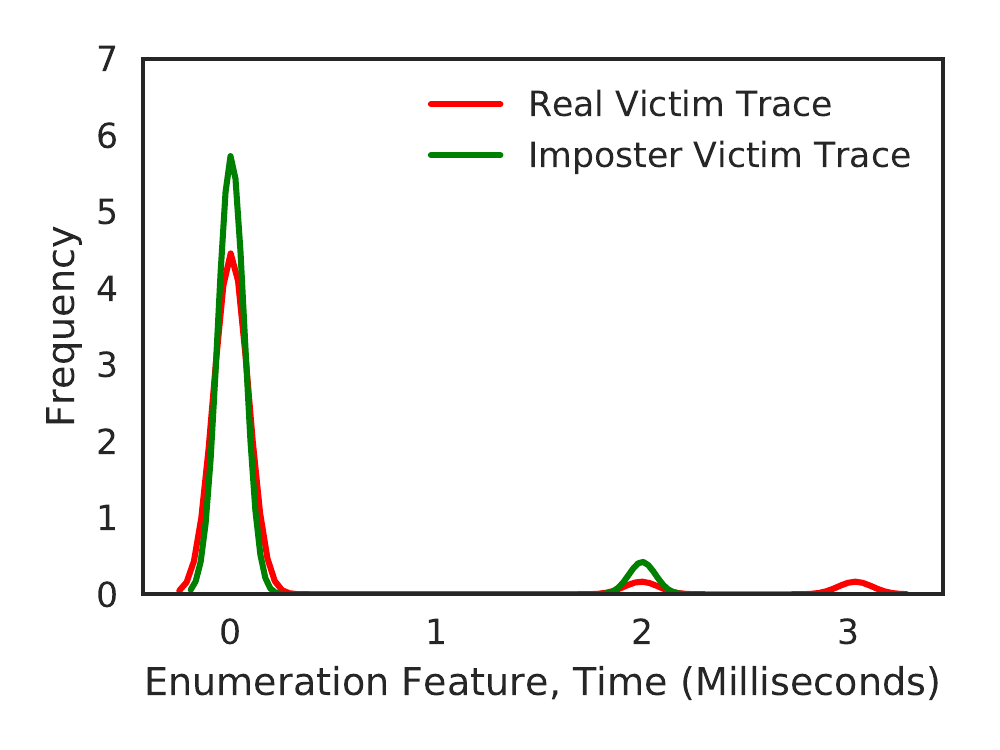}

  \caption{Comparison of distributions for a victim's real USB enumeration vector, and a faked USB enumeration vector created by the adversary.}

\label{fig:usb_dist}
\end{figure}

%% file: images/face_mats.tex
\begin{figure} 
	\centering
	\textsc{Support Vector Machine}\par\medskip

    \subfloat[][Random Pixel Image Perturb.]{\includegraphics[width=0.45\columnwidth]{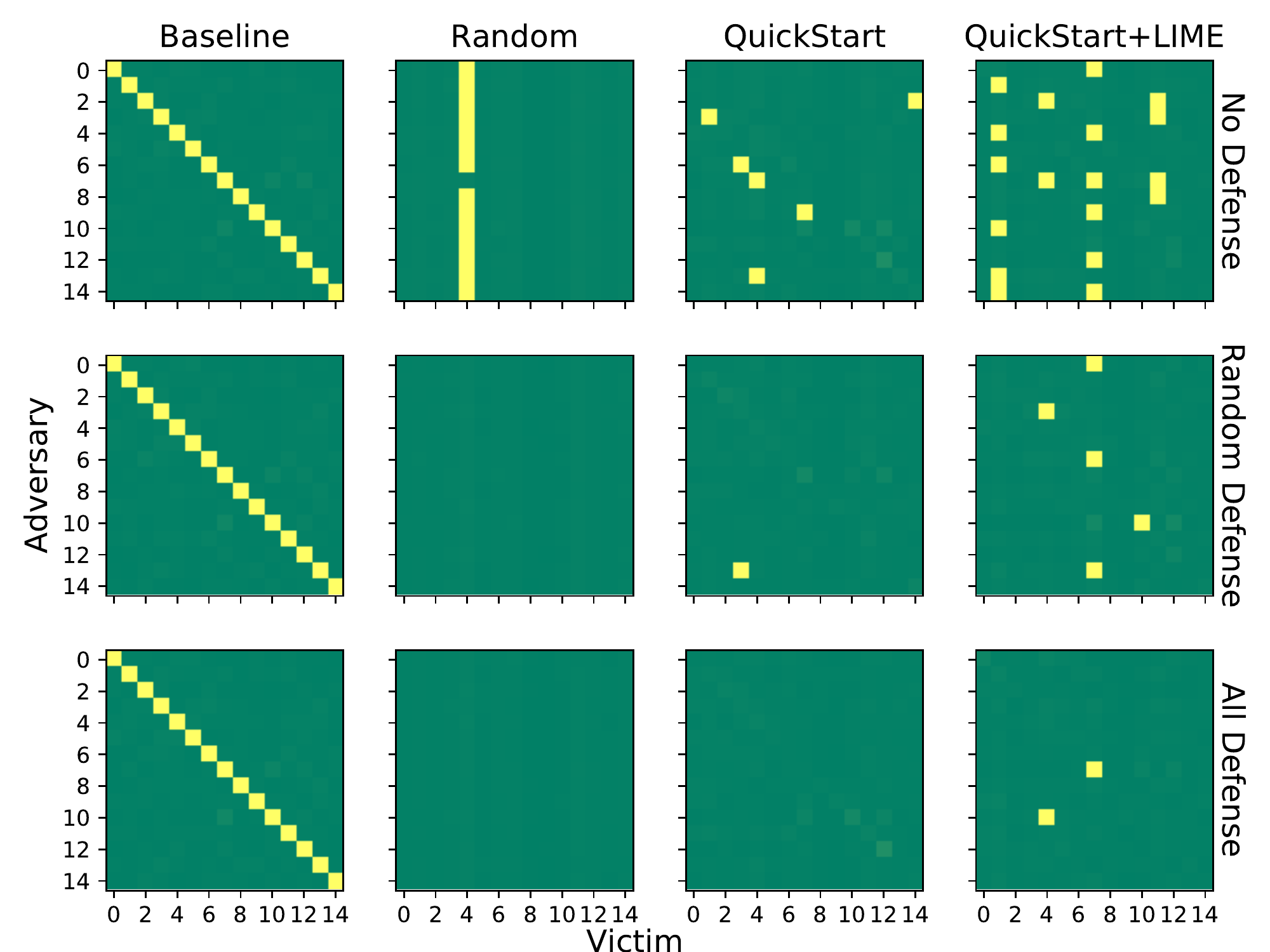}}
    \hspace{0.2cm}
    \subfloat[][VAE Latent Space Perturb.]{\includegraphics[width=0.45\columnwidth]{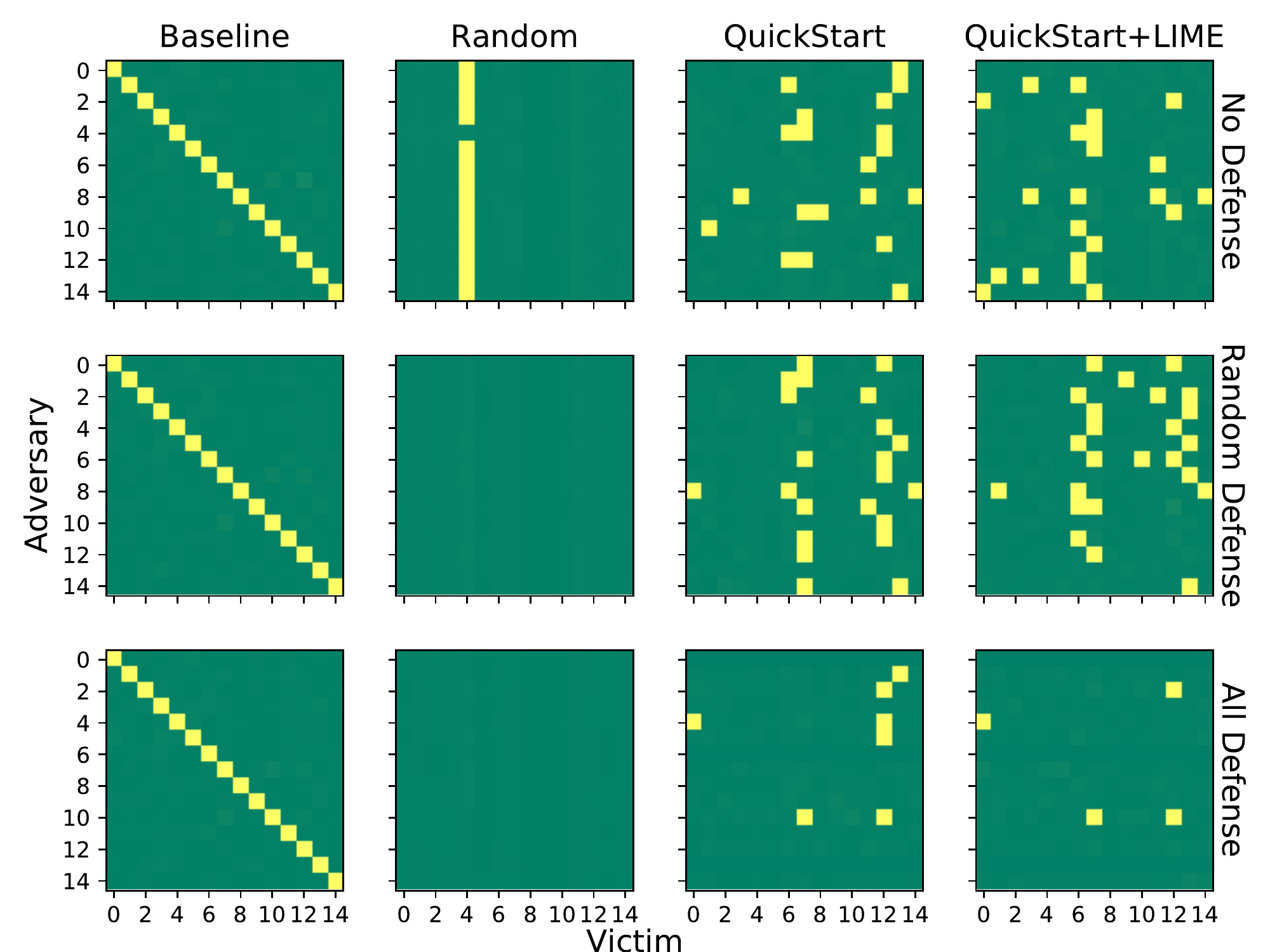}}

	\caption{Coverage maps for the Facenet-SVM classifier. Cells denote authentication success for each adversary--victim pair. Lighter cells denote unsuccessful queries shaded by the model's confidence, when available. }

\label{fig:svm_mats}
\end{figure}

%% file: images/forest_mats.tex
\begin{figure} 
  \centering

  \textsc{Random Forest}\par\medskip

  \subfloat[][Random Pixel Image Perturb.]{\includegraphics[width=0.45\columnwidth]{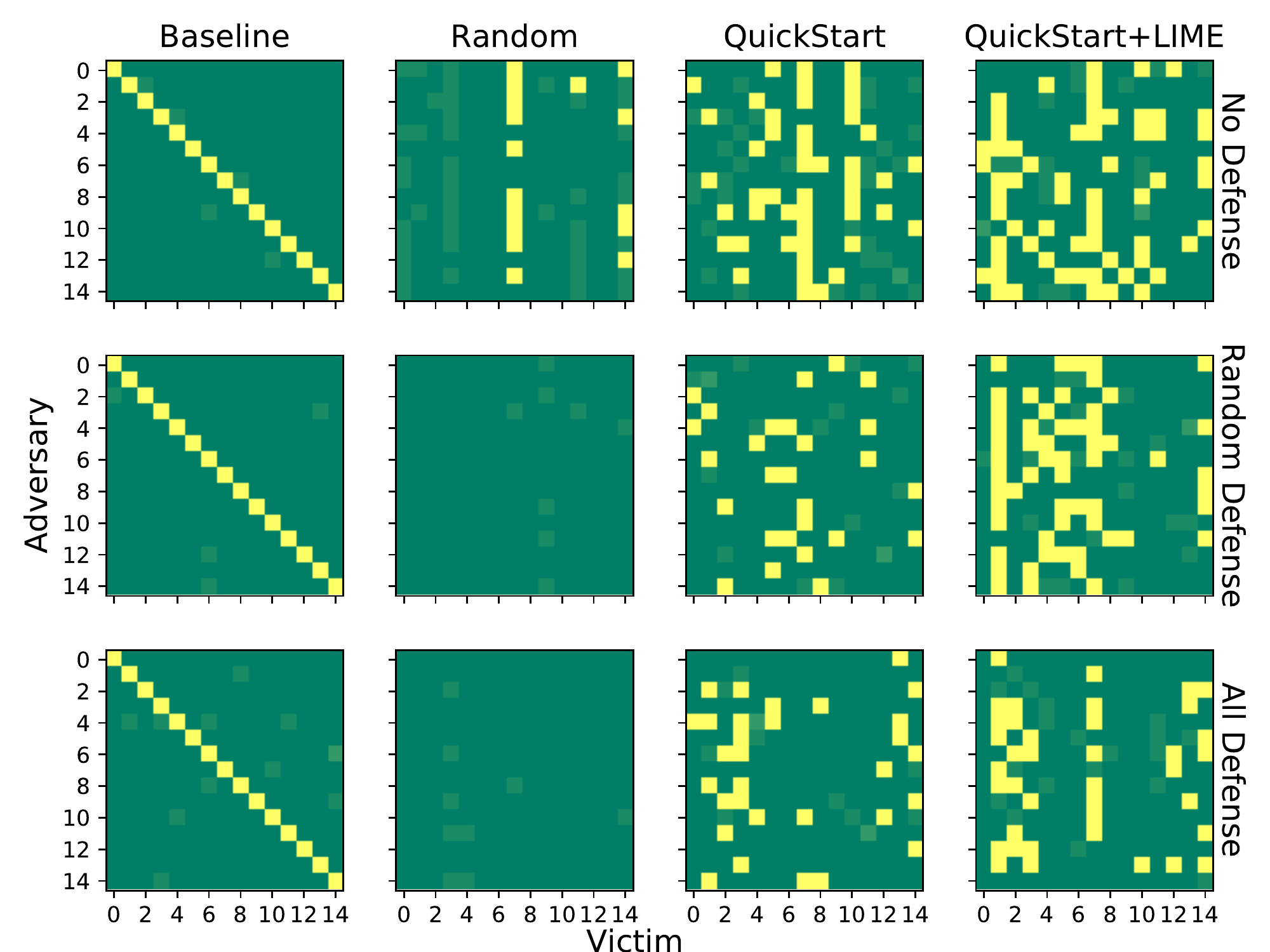}}
  \hspace{0.2cm}
  \subfloat[][VAE Latent Space Perturb.]{\includegraphics[width=0.45\columnwidth]{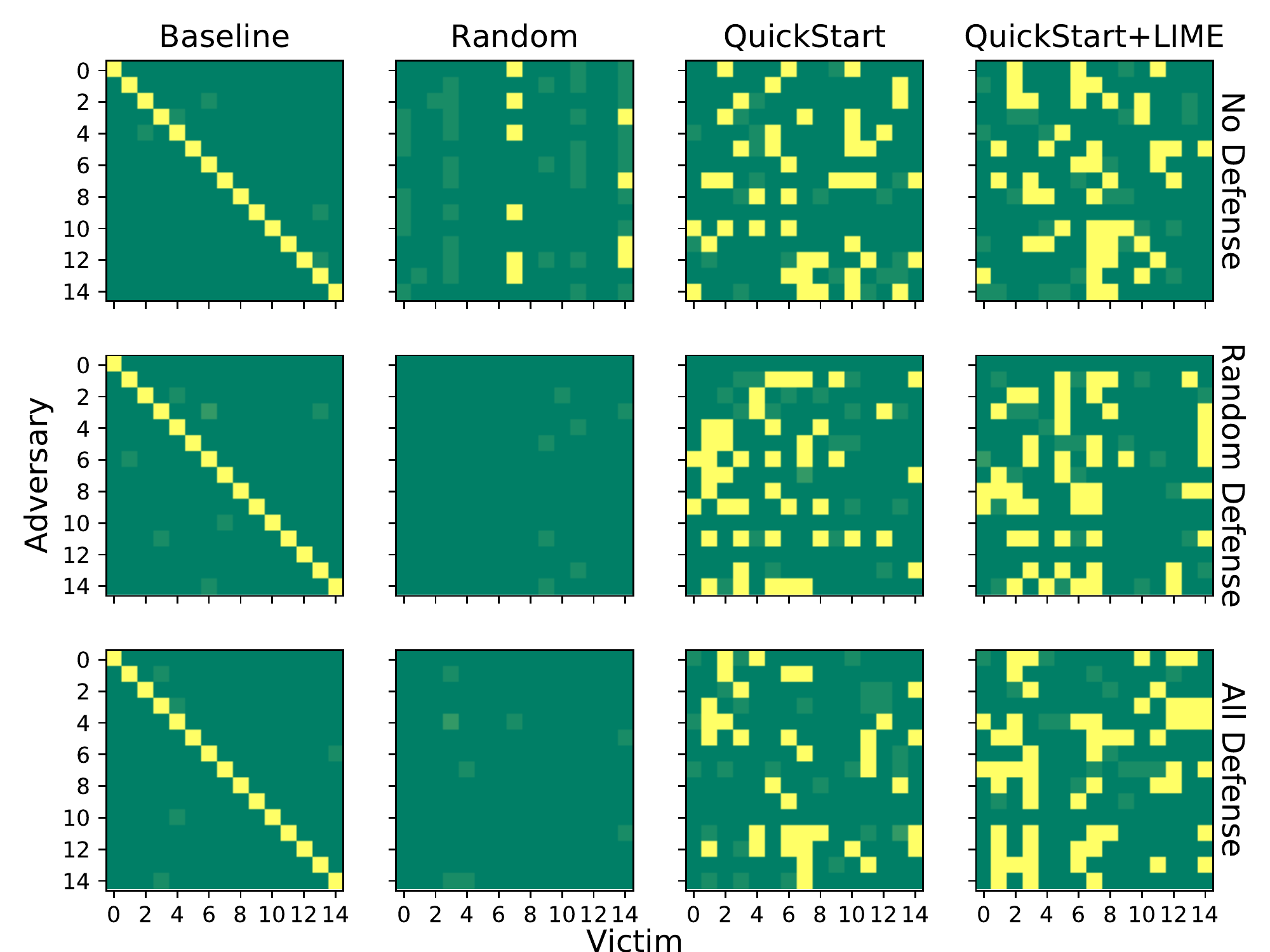}}

  \caption{Coverage maps for the Facenet-RF classifier. Cells denote authentication success for each adversary--victim pair. Lighter cells denote unsuccessful queries shaded by the model's confidence, when available. }

\label{fig:forest_mats}
\end{figure}

%% file: images/nn_mats.tex
\begin{figure} 
  \centering

  \textsc{Neural Network}\par\medskip

  \subfloat[][Random Pixel Image Perturb.]{\includegraphics[width=0.45\columnwidth]{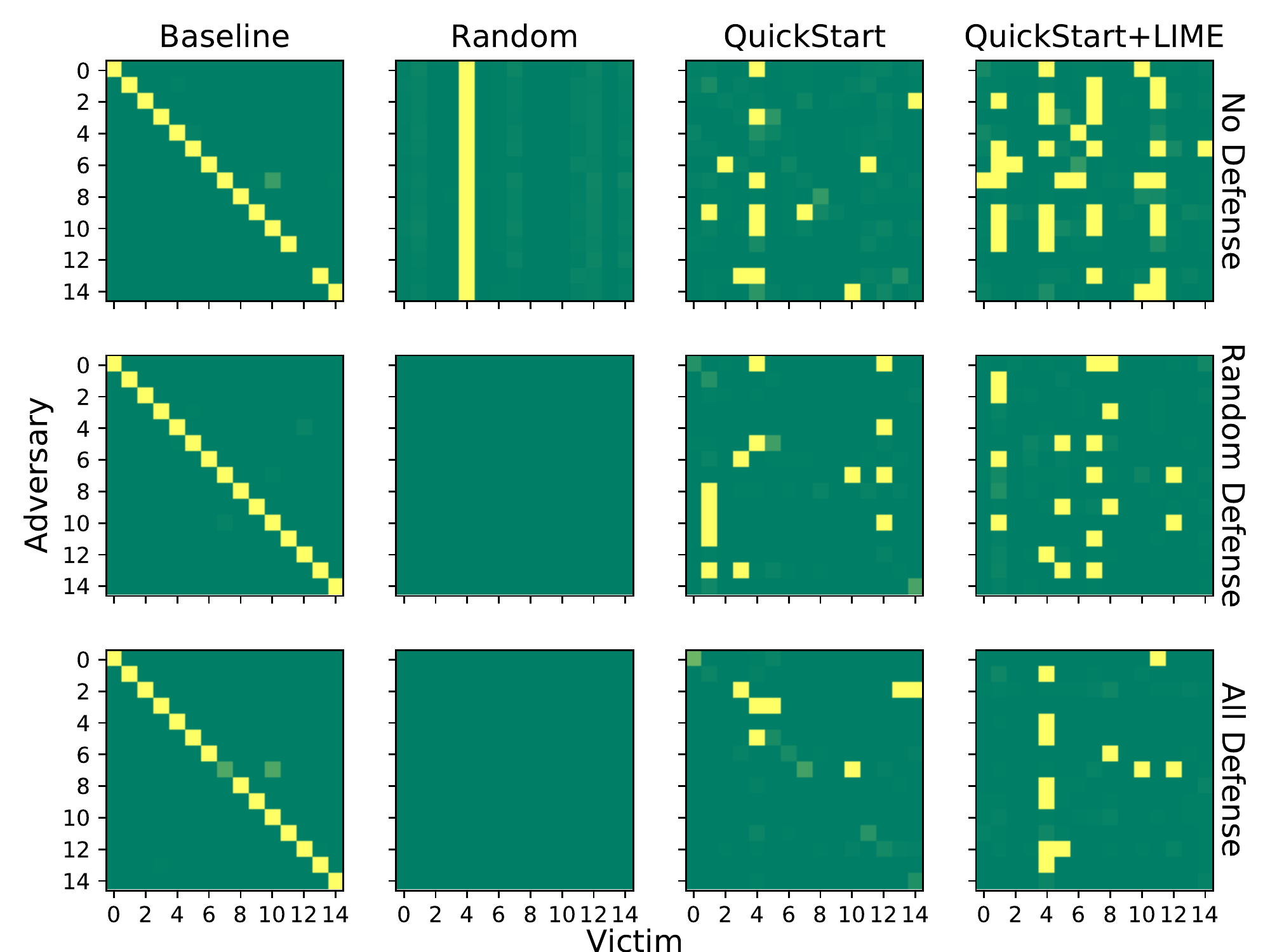}}
  \hspace{0.2cm}
  \subfloat[][VAE Latent Space Perturb.]{\includegraphics[width=0.45\columnwidth]{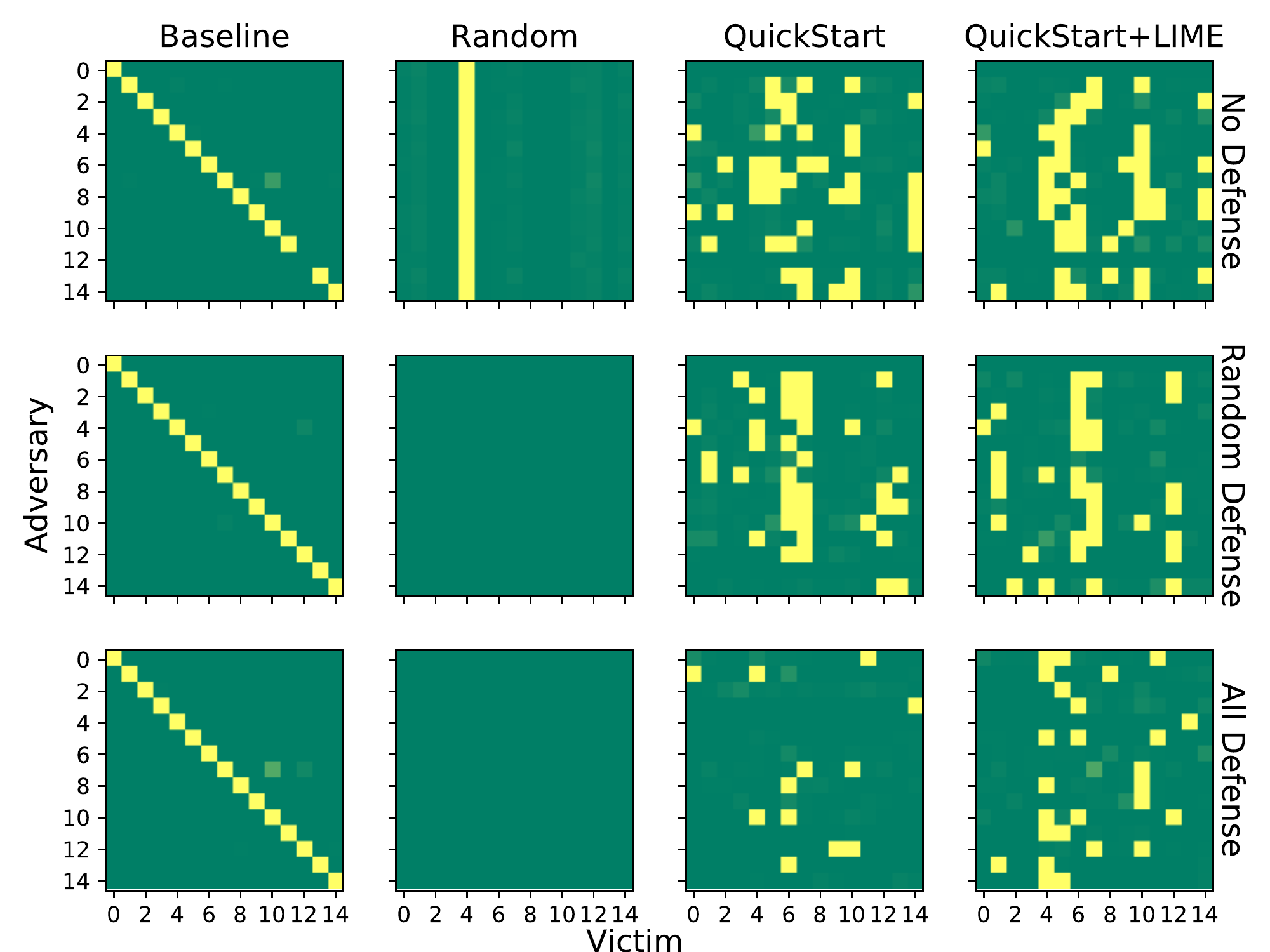}}
  
  \caption{Coverage maps for the Facenet-NN classifier. Cells denote authentication success for each adversary--victim pair. Lighter cells denote unsuccessful queries shaded by the model's confidence, when available. }

\label{fig:nn_mats}
\end{figure}

%% file: images/all_tbl.tex
\begin{table*}[t]
\begin{adjustwidth}{-.5in}{-.5in}
\tiny
\centering
\captionsetup[subfloat]{labelformat=empty}


\subfloat[]{

\begin{tabular}{|c||c|c|c|}
\multicolumn{1}{c}{} & \multicolumn{3}{c}{\normalsize{Random Pixel Image Perturb. Attack}} \\
\multicolumn{4}{c}{}\\\hline
 \textbf{SVM} & No Defense & Random Defense & All Defense\\\hline\hline
 Random & 22 $\pm$ 13.68 & ---  & --- \\
 &  $P(M)=87$\% & & \\\hline
 QuickStart & 129.5 $\pm$ 38.56 & 131 $\pm$ 0 & --- \\
 & $P(M)=40$\% & $P(M)=7$\% & \\\hline
 QuickStart & 114.67 $\pm$ 25.99 & 124.75 $\pm$ 25.91 & 152 $\pm$ 0 \\
 +LIME & $P(M)=80$\% & $P(M)=27$\% & $P(M)=7$\% \\\hline
\multicolumn{4}{c}{}\\\hline

 \textbf{RF}          & No Defense                 & Random Defense              & All Defense\\\hline\hline
 Random   &  94 $\pm$ 64.82      &  ---                         &  ---  \\  
  & $P(M)=73$\% & & \\\hline
 QuickStart &  152.47 $\pm$ 84.10 &  162.33 $\pm$ 67.12  &  107.5 $\pm$ 18.27   \\  
 & $P(M)=100$\% & $P(M)=100$\% & $P(M)=93$\% \\\hline
 QuickStart     &  120.67 $\pm$ 80.49 &  118.53 $\pm$ 26.26  &  123.71 $\pm$ 24.58   \\  
 +LIME & $P(M)=100$\% & $P(M)=100$\% & $P(M)=93$\% \\\hline
\multicolumn{4}{c}{}\\\hline
 \textbf{NN}         & No Defense                & Random Defense                & All Defense\\\hline\hline
 Random   &  13.21 $\pm$ 7.16  &   ---                         &   ---   \\  
           & $P(M)=93$\% & & \\\hline
 QuickStart &  140 $\pm$ 45.56    &   136.4 $\pm$ 53.47    &   164.75 $\pm$ 67.33   \\  
  & $P(M)=60$\% & $P(M)=67$\% & $P(M)=27$\% \\\hline
 QuickStart     &  130.23 $\pm$ 44.38 &   160.73 $\pm$  63.02  &   142.89 $\pm$ 63.58   \\  
 +LIME & $P(M)=87$\% & $P(M)=73$\% & $P(M)=60$\% \\\hline
\end{tabular}
}
%
\subfloat[]{
\begin{tabular}{|c|c|c|}
\multicolumn{3}{c}{\normalsize{VAE Latent Space Perturb. Attack}} \\
\multicolumn{3}{c}{}\\\hline
 No Defense & Random Defense & All Defense\\\hline\hline

 31.21 $\pm$ 16.42 & ---  & --- \\
 $P(M)=93$\% & & \\\hline
 102.62 $\pm$ 23.56 & 90.46 $\pm$ 12.35 & 113.6 $\pm$ 13.88 \\
 $P(M)=87$\% & $P(M)=87$\% & $P(M)=33$\% \\\hline
 100.54 $\pm$ 25.66 & 104.69 $\pm$ 22.03 & 124 $\pm$ 19.03 \\
 $P(M)=87$\% & $P(M)=87$\% & $P(M)=20$\% \\\hline
\multicolumn{3}{c}{}\\\hline
          No Defense                 & Random Defense      & All Defense\\\hline\hline
 84.11 $\pm$ 66.32  &  ---                         &  ---    \\  
 $P(M)=60$\% & & \\\hline
 125.23 $\pm$ 38.94 &  112.58 $\pm$ 50.37   &  102.43 $\pm$ 23.63   \\  
 $P(M)=87$\% & $P(M)=80$\% & $P(M)=93$\% \\\hline
 132.43 $\pm$ 35.04 &  99 $\pm$ 21.51       &  112.29 $\pm$ 39.27   \\  
 $P(M)=93$\% & $P(M)=80$\% & $P(M)=93$\% \\\hline
\multicolumn{3}{c}{}\\\hline
          No Defense                & Random Defense               & All Defense\\\hline\hline
 11.21 $\pm$ 0.41   &   ---                        &   ---   \\  
 $P(M)=93$\% & & \\\hline
 104.08 $\pm$ 25.69 &   127 $\pm$ 50.71     &   155.88 $\pm$ 46.10  \\  
 $P(M)=87$\% & $P(M)=87$\% & $P(M)=53$\% \\\hline
 108.23 $\pm$ 23.18 &   118.69 $\pm$ 29.27  &   144.43 $\pm$ 58.78  \\  
 $P(M)=87$\% & $P(M)=87$\% & $P(M)=93$\% \\\hline

\end{tabular}
}
\end{adjustwidth}
\caption{The average number of adversarial queries before success in each of three experiments, 
against each of the Facenet-SVM, Facenet-RF, and Facenet-NN classifiers and with varying level of defenses at 30\% knowledge.
$P(M)$ is the calculated probability based on attack coverage for each experiment.
Dashes indicate no successful attacks, with $P(M) = 0$.}
\label{tbl:avgqueries-all}
\end{table*}

%% file: discuss.tex
\section{Discussion}
\label{sec:discuss}







A unique benefit of using an XAI technique to produce perturbations is the ability to ``interrogate'' the substitute model $G$. By examining the explanations returned by LIME during the adversarial sample crafting process, we can form some intuition of why the attack is successful.

\subsection{Explaining Attacks}
\label{sec:exp_att}

\input{images/facepp_mattbl}


\input{images/lat_exp}

Figure~\ref{fig:lime_is_cool} illustrates an explanation of the top scoring rules in the latent variable encoding $\mathcal{L}$ for a successful attack image. We examine if LIME was accurate by perturbing one of the top scoring weights. The top scoring rule, $\mathcal{L}_{49} <= -8.16$, is the rule LIME believes to have the highest weight against the adversary. In order to force a change with only one variable, a large distortion is induced. When the variable is changed from $\mathcal{L}_{49}=2$ to $\mathcal{L}_{49}=-70$, the oracle's decision changes from $\textsc{YES}$ to $\textsc{NO}$. The new face is perceptually altered, most notably in terms of skin tone and mouth structure. This gives the adversary some insight into how the oracle is making decisions, which is a natural consequence of leveraging XAI techniques. 

\subsection{Defending with XAI}
\label{sec:defxai}

As noted previously, XAI can be used by the adversary to learn how the oracle makes decisions. However, the oracle is not restricted from using XAI on itself. The oracle may use XAI to examine which features contribute to misclassification, and determine if decisions are being made due to memorization of certain features, or by correctly extrapolating from the training data. Note however that the adversary can do the same as soon as it obtains a \textsc{YES} result. For example, the adversary is free to perform the previous experiment, and observe the effect of specific latent variables. Thus, we argue that the oracle can only win if it can generate explanations that are \textit{stronger} than those deduced by the adversary. This is not unreasonable, as the oracle has white-box access to itself. XAI systems, such as LIME~\cite{lime}, rely on iterative perturbations to discover the decision boundary. It follows that with strong explanations, an oracle can correctly infer the perturbations acting against it, and take action as needed. Future work is necessary in order to better understand the efficacy of this approach when applied as a defensive measure.

\subsection{Limitations}
\label{sec:limit}

The primary limitation in our attack is the restricted amount of data available for $G$ to learn the oracle's decision boundary. We observe this effect in LIME's performance as the feature space becomes smaller. Notably, the QuickStart+LIME attack is most effective for the Random Pixel perturbation technique, which uses the entire image as input to $G$. When training on perturbed VAE latent vectors, which are an order of magnitude smaller, the utility of LIME diminishes. Implementing data augmentation techniques into the system is an immediate improvement for future work, as it allows the adversary to further reduce the number of necessary queries. 

To the best of our knowledge, we are the first to attempt circumvention of an authentication system while assuming a blind adversary. However, due to the assumption that victim data is secret, it is not possible to target specific principals in the system. Although this is a fair assumption to make in the context of real-world authentication systems, it slightly reduces the attack's utility. However, we believe the ability to mask an adversary's malicious acts is still useful and poses a risk to any infrastructures that attempt to mediate interactions between users. 

%% file: images/facepp_mattbl.tex

\begin{figure}[t]
	\centering

    \textsc{Face++ API}\par\medskip
    \text{VAE Latent Space Perturb. Attack Coverage}\par\medskip
    \includegraphics[width=0.49\textwidth]{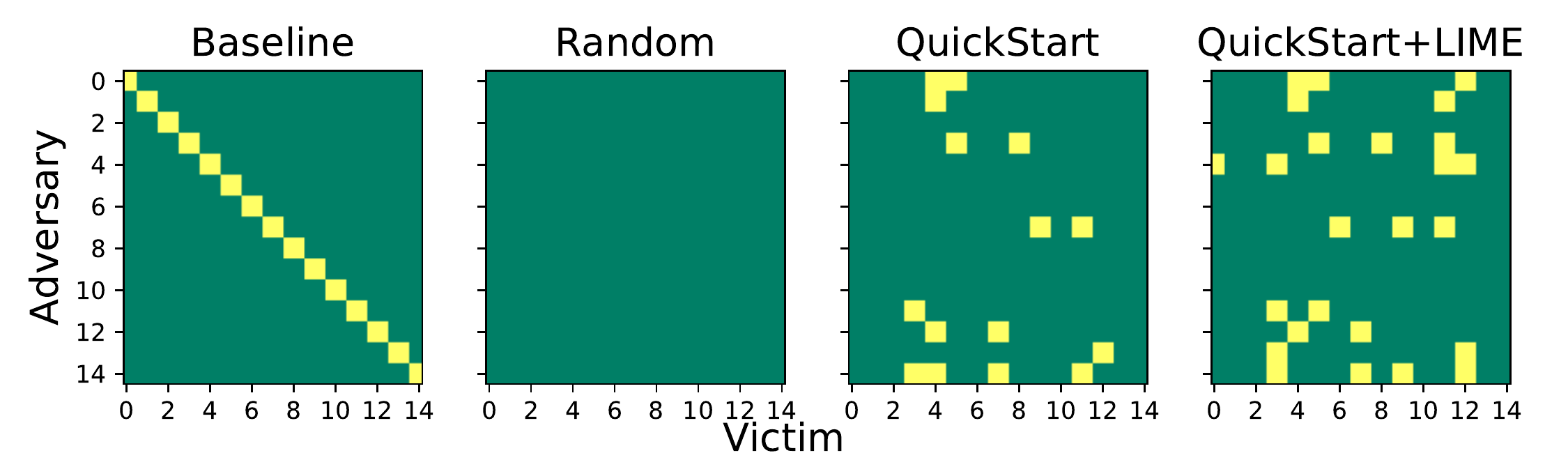}

    \captionof{figure}{Coverage heatmaps for the Face++ Black-box experiment. Cells denote authentication success for each adversary--victim pair.}
    \label{fig:facepp_mat}
    \vspace{1em}



\centering

\begin{footnotesize}

\footnotesize
\centering
\begin{tabular}{|c||c|}
\hline
 Random   & ---  \\\hline
 QuickStart & 163.88 $\pm$ 48.35\\
& $P(M)=53$\%  \\\hline
 QuickStart     & 155.11 $\pm$ 39.63\\ 
+LIME& $P(M)=60$\%  \\\hline

\end{tabular}
\end{footnotesize}
\par\medskip

\captionof{table}{The average number of adversarial queries before success in each of three experiments,
testing against Face++ API at 30\% knowledge.
$P(M)$ is the calculated probability based on attack coverage for each experiment.
Dashes indicate no successful attacks, with $P(M) = 0$.}
\label{tbl:ppqueries}


\end{figure}

%% file: images/lat_exp.tex

\begin{figure}[t]
	\centering

	\textsc{LIME Explanation}\par\medskip
	\text{Latent Variable Rules Against Adversary}\par\medskip
	\includegraphics[width=0.46\textwidth]{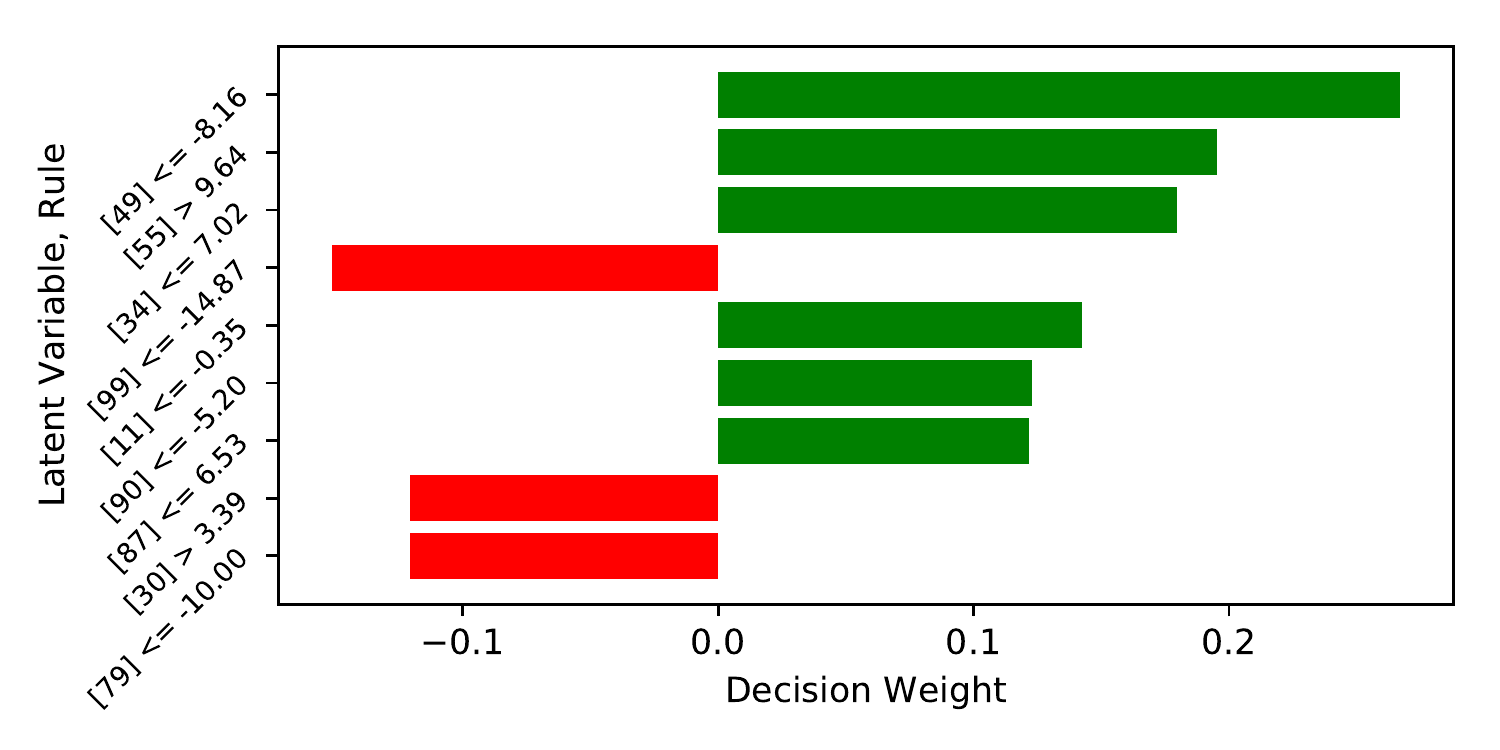}

	\includegraphics[width=0.46\textwidth]{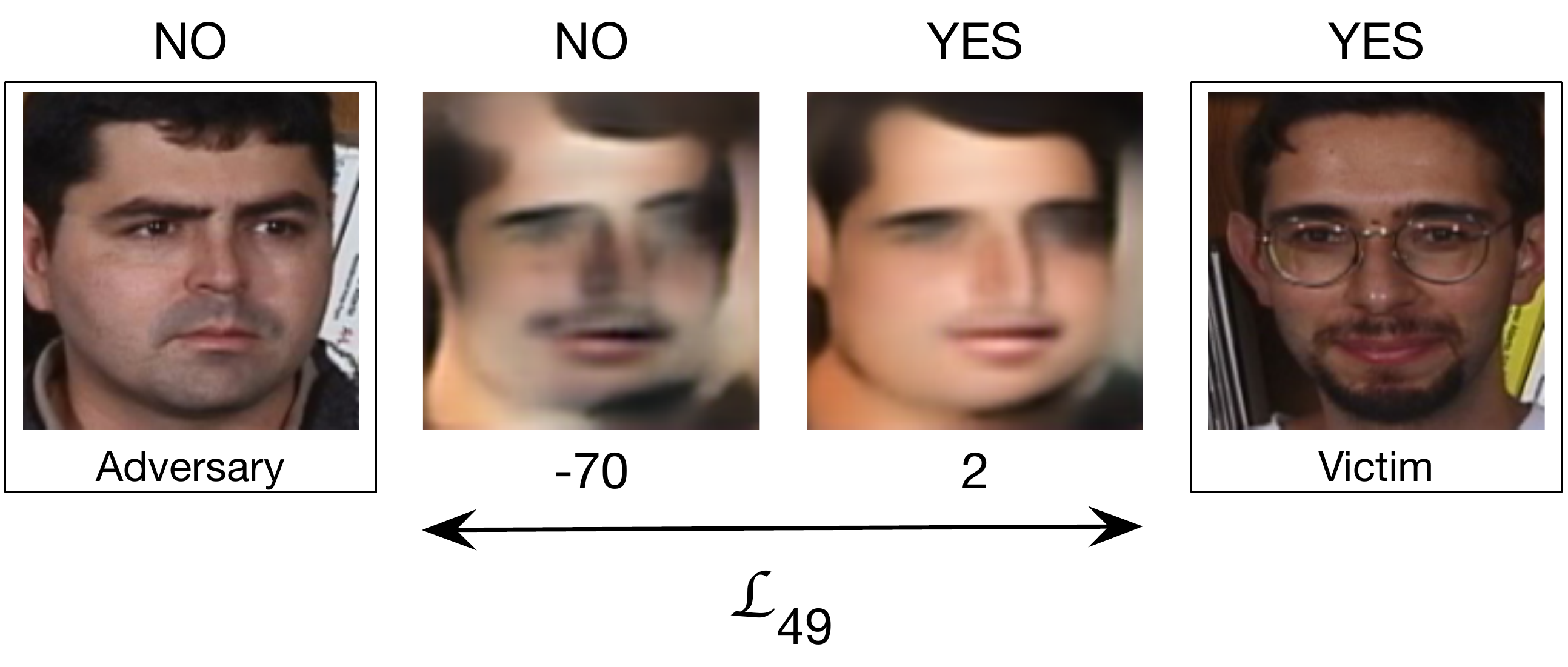}

	\caption{An explanation of potential latent variables $\mathcal{L}_i$. Rules in green denote rules towards the adversary, while rules in red denote rules away from the adversary. The most weighted latent variable, $\mathcal{L}_{49}$, is shifted to illustrate the effect on adversary's success. }

\label{fig:lime_is_cool}
\end{figure}

%% file: relwork.tex
\section{Related Work}
\label{sec:relwork}

%
Radhakrishnan et al.~\cite{Radhakrishnan2015} investigated the ability of an attacker to mimic fingerprints, but they did not consider advanced attacks that would infer the classifier's decision boundary. Such attacks were initially outlined by Barreno et al.~\cite{Barreno2006, Barreno2010} and Biggio~\cite{Biggio2014} in terms of an adversary's knowledge, intended influence, time horizon, and the type of security violation. For example, some attacks might aim to make the classifier mis-classify all future inputs, resulting in denial of service. Another attack could attempt to make gradual changes over a long period of time, successively poisoning each re-training attempt. If the fingerprinting model is rendered unusable, it must be taken offline, nullifying any security benefits it would otherwise offer. In this paper, we focus on only inferring the decision boundary at test-time, rather than trying to change it at train-time.

Machine learning classifiers designed to defend against adversaries, a technique known as adversarial machine learning, must consider attackers with varying knowledge of the classifier. Huang et al.~\cite{Huang2011} expanded on the prior work by considering particular domain applications of adversarial machine learning in both e-mail spam detection~\cite{Nelson2008} and anomalous network traffic detection~\cite{Rubinstein2009}. Attacks in these domains had to take advantage of differences in data distribution, the adversary's control over specific features, and any assumptions the learning algorithm made about the adversary. This prior work helps lay the framework for an improved fingerprinting model that may encounter adversarial inputs.

Bates et al.~\cite{Bates2014} and Radhakrishnan et al.~\cite{Radhakrishnan2015} explored several possible classifiers in their fingerprinting systems. Adversarial versions for some of these have been proposed, such as work by Dalvi~\cite{Dalvi2004} in the form of an adversary-aware Naive Bayes classifier. This technique incorporated the cost of new training samples with respect to how much it affected the decision boundary. In the realm of worm software detection, Newsome et al.~\cite{Newsome2006} proposed defenses against the Red Herring attack, where unnecessary ``chaff'' features are used to mislead the worm classifier. Zhou et al.~\cite{Zhou2012} proposed AD-SVM, an adversary-aware support vector machine (SVM), which is built against attacks that target either all possible feature modifications, or only certain data points. Although Zhou et al. show that this proposed SVM classifier is more robust, classification performance degrades when the adversary's attacks are weak. We build our own versions of these defenses to investigate their utility against a blind attacker, and compare their accuracy. Papernot et al.~\cite{Papernot2016} use previously defined adversarial sample crafting algorithms~\cite{Goodfellow2014, Papernot2015} to create an attack model capable of making deep neural network (DNN) oracles mis-classify the majority of image inputs. Unlike previous work by Barreno et al.~\cite{Barreno2006}, the target classifier is treated as a black-box model, with the only feedback being classified image labels. When an oracle also gives confidence values, Tram{\`e}r et al.~\cite{Tramer2016} show it is possible to replicate an oracle's decision tree with near-perfect accuracy. These two latter works are of key interest, as they target the final classifier techniques chosen by Bates et al. and Radhakrishnan et al., particularly the decision tree algorithm and neural networks. Although these previous attacks are useful against classification models in general, they rely on class label feedback from the model. For an authentication system, we must assume only a binary classification result, such that positive target results are secret. Regardless, we take inspiration from these previous works to build our own adversarial sample crafting algorithm. 

Modern face detection is built atop the foundational object detection framework
by Viola and Jones, which uses an ``integral image" representation to quickly
evaluate facial features~\cite{violajones}.
Face detection supports a number of applications, including face alignment,
clustering, recognition, and verification.
Specifically, we focus on facial verification systems. 
Like fingerprinting systems, facial verification
is vulnerable to targeted attacks by an adversary.
%
Fredrikson et al.~\cite{Fredrikson2015} showed that images can be recovered from API
access to facial recognition services, using only the model's output.
By contrast, the adversary in our setting perturbs its inputs without recovering the face
of target principals.
In a hill-climbing attack~\cite{hillclimbing09, hillclimbingICB}, synthetic templates are iteratively
modified based on similarity score until reaching the verification threshold, whereas our
approach does not directly iterate on feedback in this way.
Goswami et al.~\cite{GRA+18} break verification models using grid-based occlusion and face-level distortions, 
while Papernot et al.~\cite{Papernot2016} show that randomly
perturbing even a few pixels can cause misclassification. The work by Goswami et al. is the most similar to ours, as they introduce distortions or occlusions into face images to induce a misclassification. However, Goswami et al. do not consider a constrained adversary who is limited in queries. In such a case, the adversary can not only induce image processing distortions, but also construct entirely new faces using generative machine learning models, which we show is more effective (and realistic) than standard image processing distortions. 

Early XAI techniques sought to determine why opaque neural networks made
decisions by investigating their parameters and extracting rules. Towell and
Shavlik~\cite{Towell1993} extracted rules from neural networks with performance
that closely resembled that of the original model. Later,
Breiman~\cite{Breiman2001} explained random forests by randomly selecting
features and finding those with the highest influence, while Martens et
al.~\cite{Martens2007} followed a similar approach to Towell and Shavlik,
extracting rules from SVMs. These techniques eventually evolved to display more
meaningful interpretations. Although Mozina et al.~\cite{nomogram} focused on the set of the
most influential features in the model, they displayed them graphically using a
nomogram, along with a visualization of the confidence
intervals. Modern XAI techniques, such as LIME~\cite{lime},
extend this graphical approach with the use of iterative perturbations in the
style of adversarial sample crafting algorithms~\cite{Papernot2015,
  Goodfellow2014} to create human readable, globally interpretable decision
explanations. We parse these explanations to learn a concept of a potential
victim's feature space. Not only are perturbations effective, they also give
some insight into potentially vulnerable features.

To date, there has been very little investigation of XAI techniques
to address security applications. The only work we are aware of in this vein is
very recent work by Guo et al.~\cite{Guo18}, who
consider explainable techniques for deep learning-based security applications.
These types of applications, which make substantial use of RNNs, present 
challenges that are considerably different from authentication; they are focused on long
sequences, where the sequentially-oriented nature of RNNs are advantageous. By
contrast, authentication is a binary decision and lends itself well to XAI
approaches such as LIME.


%% file: conc.tex
\section{Conclusion}
\label{sec:conc}

Model-based authentication schemes are powerful methods for establishing identity in environments requiring some form of resource mediation. However, these schemes can be broken if the adversary is able to sufficiently query the oracle and learn the distinction between themselves and the other principals in the system. This paper explores the effectiveness of such an approach, and shows that feature distribution is not necessarily tied to an attacker's success. We also show that XAI is an intuitive and effective technique by which adversaries can infer decision boundaries from a victim model.

\subsection*{Acknowledgements}

We thank Nicolas Papernot for his helpful comments. This work was partially
supported by the US National Science Foundation under grant CNS-1540217 and by
ARO under contract number W911NF-1-0405.